\documentclass{article}

\usepackage[preprint]{neurips_2025}

\usepackage{wrapfig}
\usepackage[utf8]{inputenc} 
\usepackage[T1]{fontenc}    
\usepackage{hyperref}       
\usepackage{url}            
\usepackage{booktabs}       
\usepackage{amsfonts}       
\usepackage{natbib}
\usepackage{nicefrac}       
\usepackage{microtype}      
\usepackage{xcolor}         
\usepackage{graphicx}
\usepackage{amsmath}

\usepackage{algorithm}
\usepackage{algpseudocode}
\usepackage{graphicx}
\usepackage{geometry}

\usepackage{colortbl}
\usepackage{booktabs}
\definecolor{lightgreen}{RGB}{210, 255, 210}
\definecolor{darkgreen}{RGB}{0,100,0}
\definecolor{paperblue}{RGB}{83,139,230}
\definecolor{paperred}{RGB}{255,154,205}
\definecolor{paperpurple}{RGB}{186,139,222}

\usepackage{multirow}
\usepackage{makecell}

\usepackage{listings}
\definecolor{codegray}{gray}{0.95}
\definecolor{codered}{rgb}{0.8,0,0}
\definecolor{codegreen}{rgb}{0,0.6,0}
\definecolor{codeblue}{rgb}{0,0,0.7}

\lstdefinestyle{mystyle}{
    backgroundcolor=\color{codegray},   
    commentstyle=\color{codegreen},
    keywordstyle=\color{codeblue}\bfseries,
    numberstyle=\tiny\color{gray},
    stringstyle=\color{codered},
    basicstyle=\ttfamily\footnotesize,
    breaklines=true,                 
    captionpos=b,
    numbers=left,                    
    numbersep=5pt,                  
    frame=single,                   
    rulecolor=\color{black},
    showspaces=false,                
    showstringspaces=false,
    showtabs=false,
    tabsize=2
}

\usepackage{enumitem}
\setlist[itemize]{topsep=-1pt, itemsep=1pt, parsep=0pt, partopsep=0pt,left=1pt}

\algrenewcommand{\algorithmiccomment}[1]{%
  \hfill\mbox{\scriptsize\textcolor{gray}{//\,#1}}}

\title{See\&Trek: Training-Free Spatial Prompting for Multimodal Large Language Model}

\author{Pengteng Li \\ HKUST(GZ) \\ AI$^2$ROBOTICS \And Pinhao Song \\ KU Leuven \And Wuyang Li \\ EPFL \And Weiyu Guo \\ HKUST(GZ) \\ AI$^2$ROBOTICS \And Huizai Yao \\ HKUST(GZ) \And Yijie Xu \\ HKUST(GZ) \And  Dugang Liu \\SZU \And Hui Xiong \\ HKUST(GZ) }

\begin{document}

\maketitle

\begin{abstract}
We introduce \textbf{\textsc{See\&Trek}}, the first training-free prompting framework tailored to enhance the spatial understanding of Multimodal Large Language Models (\textsc{MLLMs}) under vision-only constraints. While prior efforts have incorporated modalities like depth or point clouds to improve spatial reasoning, purely visual-spatial understanding remains underexplored. \textbf{\textsc{See\&Trek}} addresses this gap by focusing on two core principles: increasing visual diversity and motion reconstruction. For visual diversity, we conduct Maximum Semantic Richness Sampling, which employs an off-the-shell perception model to extract semantically rich keyframes that capture scene structure. For motion reconstruction, we simulate visual trajectories and encode relative spatial positions into keyframes to preserve both spatial relations and temporal coherence. Our method is training\&GPU-free, requiring only a single forward pass, and can be seamlessly integrated into existing \textsc{MLLMs}. Extensive experiments on the \textsc{VSI-Bench} and \textsc{STI-Bench} show that \textbf{\textsc{See\&Trek}} consistently boosts various \textsc{MLLMs} performance across diverse spatial reasoning tasks with the most +3.5\% improvement, offering a promising path toward stronger spatial intelligence.
\end{abstract}

\vspace{-1em}
\section{Introduction}
\vspace{-0.5em}

Multimodal Large Language Models (\textsc{MLLMs})~\cite{qwen25,DeepseekR1,gpt,team2024gemini} have witnessed rapid advancements, demonstrating impressive capabilities in understanding and generating cross-modal content. By integrating visual and textual inputs, these models have shown potential in various tasks such as image captioning~\cite{VideoMMMU}, visual question answering~\cite{LLaVA-NeXT}, and embodied AI~\cite{liu2024nvila,Openvla,LLaKey}. 

Spatial reasoning is crucial for empowering MLLMs to understand and interact with real-world environments, particularly in tasks involving object localization, motion prediction, and physical interactions. Enhancing MLLMs with spatial awareness can significantly improve their performance on downstream applications such as navigation~\cite{Drivegpt4} and robotic manipulation~\cite{Openvla,CotVla}. Recent works have introduced depth cues~\cite{gpt}, camera poses~\cite{llava3d}, and 3D priors~\cite{SpatialRGPT} into MLLM training, aiming to construct more comprehensive spatial representations~\cite{Scanqa,Sqa3d}. However, despite these advances, existing models still struggle to robustly reason about complex spatial relationships, particularly in scenarios requiring multi-step reasoning, complex visual and temporal dynamics, or generalization to novel environments. \textit{We reflect critically on two pivotal factors that hinder current \textsc{MLLMs} from overcoming the spatial understanding bottleneck:}

\begin{figure}
    \centering
    \includegraphics[width=\linewidth]{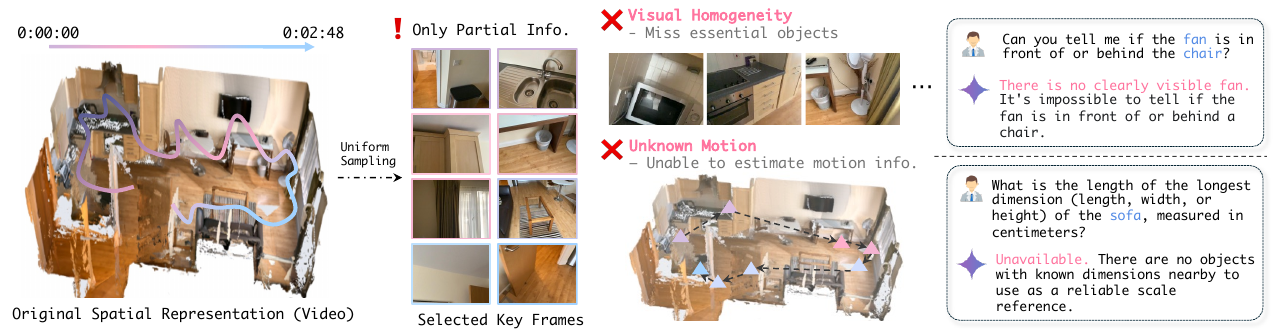}
    \vspace{-0.4cm}
    \caption{Illustration of insufficient spatial understanding ability of current \textsc{MLLMs}. They adopt uniform sampling to extract the video with spatial semantics, leading to \textcolor{paperred}{visual homogeneity} and \textcolor{paperred}{unknown motion.} \textbf{\textsc{See\&Trek}} aims to tackle these two problem, which can be referred to Figure \ref{fig:results}.}
    \label{fig:abstract}
    \vspace{-1.5em}
\end{figure}

\noindent \textbf{1) Visual Homogeneity}: As shown in Figure \ref{fig:abstract}, generally, a large number of existing \textsc{MLLM} pipelines adopt uniform temporal sampling strategies, in which 8 or 32 frames are used as input due to limited GPU memory when selecting keyframes from the video which captures full spatial semantics.  Without any structural constraints or prior knowledge, this uniform temporal sampling often captures the frames (i) without any salient features (e.g., walls, ceilings, and floors) (ii) or containing fragments of objects. Those frames will decrease the signal-to-noise ratio in the input frames, limiting the MLLM’s ability to reconstruct or reason about the full spatial layout.

\noindent \textbf{2) Unknown Motion}: Relying solely on sampled frames, without access to explicit ego-motion information, significantly impairs a model’s ability to infer object movement and displacement within a scene. Such capabilities are critical to spatial reasoning tasks, including estimating object distances, predicting motion trajectories, and establishing temporal order. In the absence of explicit motion cues, MLLMs are forced to rely primarily on commonsense priors acquired during pretraining, rather than on directly grounded visual evidence. Consequently, the spatial predictions made by such models tend to be speculative rather than evidence-based, revealing a fundamental limitation of existing vision-only MLLM architectures.

To tackle the two above-mentioned important factors, we propose \textbf{\textsc{See\&Trek}}, a simple yet effective training\&GPU-free spatial prompt method to jointly boost spatial-temporal reasoning in \textsc{MLLMs}. \textit{To tackle the visual homogeneity,} we utilize an off-the-shelf perception model, e.g., object detectors, to extract semantically rich keyframes that capture the spatial structure of scenes to increase visual diversity. \textit{To tackle unknown motion,} we leverage Visual Odometry (VO) to simulate visual trajectories from the given videos and add extra motion cues into the keyframes, preserving spatial relationships and temporal coherence. Featured with 1) training\&GPU-free, 2) plug-and-play, and 3) single-forward characteristics, \textbf{\textsc{See\&Trek}} can be seamlessly integrated with open-source \textsc{MLLMs} or commercial engines. 
Comprehensive experiments on \textsc{VSI-Bench} and \textsc{STI-Bench} verify that \textbf{\textsc{See\&Trek}} significantly improves the performance of multiple \textsc{MLLMs} across diverse spatial reasoning tasks, offering a promising direction for enhancing the spatial understanding of \textsc{MLLMs}. In summary, our contributions are as follows:
\begin{itemize}
    \item We introduce \textbf{\textsc{See\&Trek}}, the first \textbf{training- and GPU-free} spatial prompting framework to enhance the spatial understanding capabilities of \textsc{MLLMs}. It is \textbf{plug-and-play}, \textbf{single-forward}, and compatible with both open-source and commercial MLLMs.
    \item To provide rich spatial and motion cues, we design a comprehensive spatial prompting strategy: 1) Maximum Semantic Richness Sampling: We use off-the-shelf perception models (\textit{e.g.}, \textsc{YOLO}) to extract semantically rich and diverse keyframes from videos. 2) Motion Reconstruction: We reconstruct camera motion in BEV and 3D using Visual Odometry and label keyframes with motion cues to preserve spatial and temporal coherence.
    \item Comprehensive experiments on \textsc{VSI-Bench} and \textsc{STI-Bench} show that \textbf{\textsc{See\&Trek}} significantly boosts the spatial understanding performance of multiple \textsc{MLLMs} across diverse scales and architectures, offering a promising direction for future spatial intelligence.
\end{itemize}

\vspace{-0.75em}
\section{Related Works}
\vspace{-0.5em}

\textbf{Long Video Understanding.} Most evaluations of spatial understanding in \textsc{MLLMs} focus on long video comprehension~\cite{VideoMMMU,Mlvu}, where models analyze dynamic scenes and answer related questions. A key challenge in both spatial and video understanding is maximizing relevant information retention. Existing methods often employ specialized adapters~\cite{HierarQ,breakingencoder}, tokenization strategies~\cite{QuoTA,hybridtoken}, or memory modules~\cite{yuan2025memory,infty-Video,ReWind} to compress and store video semantics. Other approaches convert videos into document-style inputs~\cite{ma2024drvideo}, use retrieval-augmented generation (RAG) frameworks~\cite{VideoRAG-l,VideoRAG-s,Graphsvideo,Videotree}, or structure content in trees~\cite{Videotree} to reduce memory load. However, these techniques generally require fine-tuning \textsc{MLLMs} or rely on \textsc{VLMs} for information retrieval, which is both computationally expensive and resource-intensive. Recent work has explored query-based frame selection~\cite{adaptiveframes,agentic-keyframes,GenerativeFrameSampler,LogicinFrames,Selection}, combining traditional heuristics with feedback from pretrained \textsc{VLMs}. Building on these insights, we propose a novel GPU-free keyframe selection strategy that identifies representative frames efficiently without additional training or heavy computational demands.

\textbf{Spatial Understanding.} Spatial understanding is essential for deploying \textsc{MLLMs} in real-world applications such as Embodied AI~\cite{Openvla,CotVla} and Autonomous Driving~\cite{drivesurvey,Drivegpt4}. Recent research has explored leveraging multimodal inputs, such as depth maps or point clouds, to enable explicit~\cite{GPT4Scene} or implicit~\cite{llava3d,STVLM,SpatialRGPT} 3D scene modeling, which is processed by the LLM decoder to enhance spatial comprehension. However, these approaches often rely on precise cross-modal alignment and highly customized pipelines, posing challenges for deployment and limiting generalizability across diverse real-world settings. Additionally, common benchmarks like ScanQA~\cite{Scanqa} and SQA3D~\cite{Sqa3d} typically assume uniform camera motion, making them less suitable for complex, dynamic environments. In contrast, vision-only \textsc{MLLMs} offers greater usability~\cite{qwen25,DeepseekR1,team2024gemini} but often struggles with nuanced spatial understanding in video-based tasks~\cite{kimi,vsibench,STIBench}. These limitations motivate the development of a spatial prompting strategy that enhances the spatial reasoning capabilities of existing \textsc{MLLMs}.

\begin{figure}[t]
    \centering
    \includegraphics[width=\linewidth]{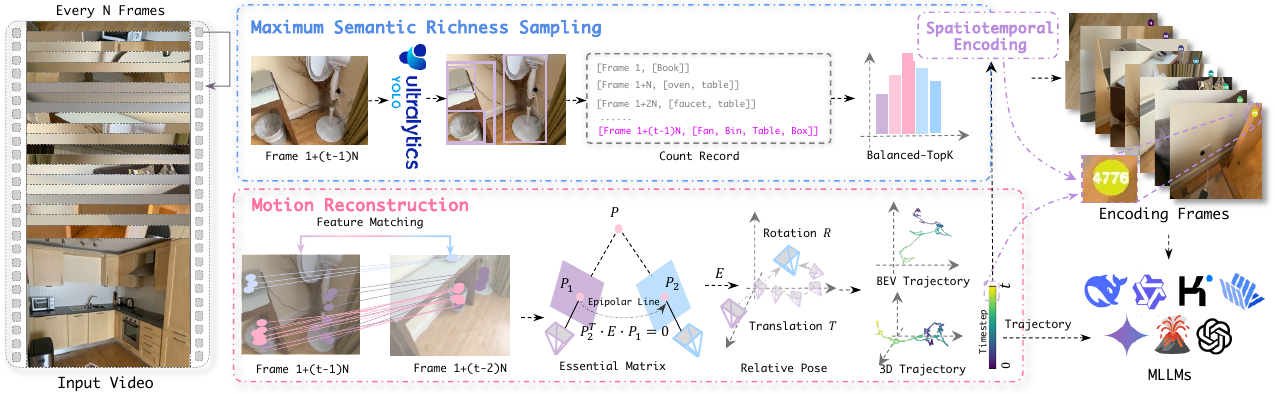}
    \vspace{-0.4cm}
    \caption{Overview of the proposed \textbf{\textsc{See\&Trek}}. We first sample one frame for every $N$ frame for post-processing.  For \textcolor{paperblue}{Maximum Semantic Richness Sampling}, a visual perception model \textit{e.g.} \textsc{YOLO}~\cite{ultralytics2023yolov8} detects objects in each sampled frame, updating the category count. To boost scene representation, we propose a \textit{Balanced-Top$K$} strategy for selecting semantically rich frames. For \textcolor{paperred}{Motion Reconstruction}, we perform feature matching with the previous frame. Then, matching points are used to estimate the essential matrix $E$ via RANSAC~\cite{RANSAC}, from which the relative camera pose and motion trajectory are recovered. Finally, \textsc{Spatiotemporal Encoding} is introduced to integrate motion and spatial cues for a more comprehensive representation.}
    \vspace{-1.5em}
    \label{fig:framework}
\end{figure}

\vspace{-0.75em}
\section{Method}
\vspace{-0.5em}

As shown in Figure \ref{fig:framework}, the proposed \textbf{\textsc{See\&Trek}} aims to overcome the limitation of visual homogeneity and unknown motion in the captured keyframes.
In detailed, given the video sequence $V = \{f_t\}^{N_v}_{t=1}$ with ${N_v}$ frames, we sample one frame every $N$ frames of $V$. We first select the rich semantics frames by leveraging an efficient visual perception model and the proposed \textit{Balanced-Top$K$} strategy and use OV to simulate the camera relative motion trajectory to obtain the moving information. Finally, \textsc{Spatiotemporal Encoding} is introduced to encode each frame with color-specified and number marks, which efficiently integrates both important properties. The algorithm is presented in Algorithm~\ref{alg:main} and the detailed version can be found in Algorithm~\ref{alg:detailed} in the appendix.

\begin{algorithm}[t]\small
\caption{\textsc{See\&Trek}: Maximum Semantic Richness Sampling \& Motion Reconstruction}
\label{alg:main}
\begin{algorithmic}[1]
\Require Video $V=\{f_t\}_{1}^{N_v}$, interval $N$, \#keyframes $K$, detector $\mathcal{Y}$, intrinsics $\mathbf{K}$
\Ensure Keyframes $\{\mathcal{F}'_{t_i}\}_{i=1}^{K}$, $\mathbf{P}_{\text{BEV}}$, $\mathbf{P}_{\text{3D}}$, $\mathcal{T}^{\text{world}}$

\State $\mathcal{S}\gets\{f_t\mid t\bmod N=0\}$ \Comment{Uniform Subsampling}
\State $\mathcal{T}^{\text{world}}\gets[]$, $\mathcal{O}\gets\emptyset$
\For{each pair $(f_{t-1},f_t)\in\mathcal{S}$}
    \State $(\mathbf{R}_t,\mathbf{T}_t,\mathcal{C}_t)\gets\textsc{FramePipeline}(f_{t-1},f_t)$%
          \Comment{ORB$\to$Match$\to$RANSAC$\to$Decompose \& Perception $\mathcal{Y}$}
    \State $\mathcal{T}^{\text{world}}{+}{=}\mathbf{T}_t^{\text{world}},\;
           \mathcal{O}{+}{=}(t,\mathcal{C}_t)$
\EndFor

\State $\mathcal{O}^{*}\gets\textsc{TrimSpan}(\mathcal{O})$ \Comment{Valid Interval $[t_s,t_e]$}
\State $(\mathcal{T}_{\text{sel}},\mathcal{C}_{\text{sel}})\gets
       \textsc{SelectKeyframes}(\mathcal{O}^{*},K)$%
       \Comment{\textit{Balanced‐Top$K$}}
\For{$t_i\in\mathcal{T}_{\text{sel}}$}
    \State $\mathcal{F}'_{t_i}\gets\textsc{Overlay}(f_{t_i},\text{ColorMap}(t_i/K))$ \Comment{\textsc{Spatiotemporal Encoding:} Index \& Hue Mark}
\EndFor
\State $(\mathbf{P}_{\text{BEV}},\mathbf{P}_{\text{3D}})\gets
       \textsc{RenderTraj}(\mathcal{T}^{\text{world}})$ \Comment{ProjectXY + Render3D}
\end{algorithmic}
\end{algorithm}
\vspace{-0.5em}


\vspace{-0.5em}
\subsection{Maximum Semantic Richness Sampling}
\vspace{-0.5em}
\label{sec:Spatial Visibility}

To address the limitation of \textbf{visual homogeneity}, the first step of \textbf{\textsc{See\&Trek}} is to select the keyframes that convey the richest semantic content to increase visual diversity. Since inference resources constrain us to only a few frames, we posit that the frame containing the most detected objects will best capture the scene’s diversity. Concretely, we run a pretrained YOLO detector $\mathcal{Y}(\cdot)$ ~\cite{ultralytics2023yolov8} on every frame $f_t$, yielding a detected class set $\mathcal{C}_t = \mathcal{Y}(f_t) = \{c_1, c_2, \dots, c_{n_t}\}$, 
where $n_t$ is the number of detected classes, and each $c_i$ belongs to the predefined label vocabulary $\mathcal{L}$. Thus, within a valid temporal interval $[t_s, t_e]$, we collect a set of detected class set for each frame $f_t$ $\mathcal{C}^* = \left\{  \mathcal{C}_{t} \,\middle|\, t_s \leq t \leq t_e \right\}.$

A simple frame selection method is to choose the Top$K$ frames with the most detected classes. However, this can bias selection toward the temporal interval in which the greatest number of objects appear, causing MLLMs to miss other semantically important parts of the video. To address this, we propose a frame sampling strategy called \textit{Balanced-Top$K$}, which extracts a set of \(K\) keyframes that are both object-rich and temporally diverse.
In detail, we first select the initial keyframe\(f_{\tau_0}\) as the one with the maximum number of detected classes:
\begin{equation}
    \tau_g = \arg\max_{t_s \leq t \leq t_e} \left( |\mathcal{C}_{t}| \right),
\end{equation}
with ties broken by choosing the earliest frame. The initial class pool is set as $\mathcal{C}_{\text{sel}} = \mathcal{C}_{t_0}$. 
To avoid local bias, ensure temporal uniformity, and capture a broader semantic representation, we partition $\mathcal{C}^*$ into $K-1$ contiguous temporal segments $\left\{ \mathcal{C}^{(1)}, \dots, \mathcal{C}^{(K-1)} \right\}$. In each segment $\mathcal{C}^{(k)}$, to increase the overall information gain per frame, we select one frame $\tau_k$ such that the overlap of its detected categories with $\mathcal{C}_{\text{sel}}$ is minimized, classes count is maximized, and the frame is as early as possible:
\begin{equation}
    \tau_k = \arg\min_{t \in \mathcal{T}^{(k)}} \left(
        |\mathcal{C}_{t} \cap \mathcal{C}_{\text{sel}}|,\;
        -|\mathcal{C}_{t}|,\;
        t
    \right),
\end{equation}
where $\mathcal{T}^{(k)}$ is the subset of timesteps, in which each timestep $\tau$ corresponds to a detected class set $\mathcal{C}_\tau$ in $\mathcal{C}^{(k)}$.
After each selection, the accumulated category set is updated as $\mathcal{C}_{\text{sel}} \leftarrow \mathcal{C}_{\text{sel}} \cup \mathcal{C}_{\tau_k}.$
This process yields a set of $K$ selected keyframes $\{\tau_i\}_{i=0}^{K-1}$ that are semantically representative, object-abundant, and temporally distributed, forming an optimal subset for downstream understanding.

\vspace{-0.5em}
\subsection{Motion Reconstruction}
\vspace{-0.5em}
\label{sec:Motion Reconstruction}
As mentioned in the introduction, MLLMs' spatial understanding is limited by \textbf{unknown camera motion}, which leads to confusion about movement patterns, object displacement, and the spatial layout of a scene. In this section, we aim to estimate the camera motion via visual odometry (VO) from a monocular video and connect the estimated camera motion with the selected keyframes by painting motion cues on them. Through feeding MLLMs with visualized camera motion and keyframes with motion cues, the spatial understanding capability can be enhanced.

\noindent \textbf{Camera motion estimation.} We adopt a feature-based VO pipeline utilizing ORB (Oriented FAST and Rotated BRIEF) features and essential matrix estimation~\cite{OrbSlam3,OrbSlam}. In detail, the current frame $f_t$ and the last frame $f_{t-1}$ are converted to grayscale. We first extract ORB key points and descriptors from each frame as follows:
\begin{equation}
\begin{cases}
   \mathcal{K}_{t-1}, \mathcal{D}_{t-1} &= \text{ORB}(f_{t-1}), \\
   \mathcal{K}_t, \mathcal{D}_t &= \text{ORB}(f_t),
\end{cases}
\end{equation}
where $\mathcal{K}_{t-1} = \{ \mathbf{p}_i^{t-1} \}_{i=1}^{N}$ and $\mathcal{K}_t = \{ \mathbf{p}_i^t \}_{i=1}^{N}$ denote the detected 2D keypoints (in pixel coordinates), and $\mathcal{D}$ represents the corresponding ORB binary descriptors. Feature correspondences are established by matching descriptors between the two frames, resulting in a set of matched keypoint pairs $
    \mathcal{M}_{t-1,t} = \left\{ (\mathbf{x}_i^{t-1}, \mathbf{x}_i^t) \right\}_{i=1}^{P},$
where $\mathbf{x}_i^{t-1}, \mathbf{x}_i^t \in \mathbb{R}^2$ are the matched 2D keypoints in pixel coordinates, and $P$ is the number of matches. To compute the essential matrix, we first normalize the image coordinates using the camera intrinsic matrix $\mathbf{K}$:$
\hat{\mathbf{x}}_i = \mathbf{K}^{-1} \begin{bmatrix} \mathbf{x}_i \\ 1 \end{bmatrix}.$
The essential matrix $\mathbf{E} \in \mathbb{R}^{3 \times 3}$ encapsulates the relative motion between the two frames (up to scale) and is estimated by minimizing the geometric error of the epipolar constraint:
\begin{equation}
    \mathbf{E} = \arg\min_{\mathbf{E'}} \sum_i \rho\left( (\hat{\mathbf{x}}_i^t)^\top \mathbf{E'} \hat{\mathbf{x}}_i^{t-1} \right),
\end{equation}
where $\rho(\cdot)$ is a robust loss function such as the truncated quadratic or Tukey biweight~\cite{huber2011robust}, and RANSAC is used to handle outliers~\cite{RANSAC}. Once the essential matrix $\mathbf{E}$ is estimated, we recover the relative rotation $\mathbf{R}_t \in SO(3)$ and translation direction $\mathbf{T}_t \in \mathbb{R}^3$ (up to scale) by decomposing $\mathbf{E}$ via Singular Value Decomposition (SVD) and applying the Cheirality condition to resolve ambiguity. This results in the relative pose $(\mathbf{R}_t, \mathbf{T}_t)$ between frames $t-1$ and $t$. Assuming the global camera pose at time $t-1$ is known as $(\mathbf{R}_{t-1}^{\text{world}}, \mathbf{T}_{t-1}^{\text{world}})$, the global pose at time $t$ is updated recursively:
\begin{equation}
\begin{cases}
    \mathbf{R}_t^{\text{world}} = \mathbf{R}_{t-1}^{\text{world}} \mathbf{R}_t, & \\
    \mathbf{T}_t^{\text{world}} = \mathbf{R}_{t-1}^{\text{world}} \mathbf{T}_t + \mathbf{T}_{t-1}^{\text{world}}. &
\end{cases}
\end{equation}
This process is repeated over the entire sampled frame sequence to reconstruct the full camera trajectory. By estimating relative camera motion, we obtain the spatial position of each frame and the overall structure of the scene, thereby offering an implicit spatial reference for the orientation and placement of objects within the space. After obtaining the 3D camera trajectory $\{ \mathbf{T}_t^{\text{world}} \}_{t=1}^{K}$, we generate two visualizations: a top-down Bird’s Eye View (BEV) and a 3D trajectory plot, denoted as $\mathbf{P}_{\text{BEV}}$ and $\mathbf{P}_{\text{3D}}$. For visualization, each trajectory point at timestamp $t$ is assigned a color using $\Phi(t) = \text{ColorMap}(t / K),$ where $K$ denotes the total number of trajectory points. The BEV is created by projecting the 3D camera positions onto the XY plane, while the 3D plot retains the full spatial geometry. These visualizations provide a concise and intuitive summary of the camera motion throughout the video.

\textbf{Spatiotemporal Encoding.} Though the semantically rich keyframes and camera motion are obtained, MLLMs have difficulty in connecting both information if they are simply fed into MLLMs: \textit{MLLMs have no idea of the camera poses of the corresponding keyframes.} This disconnection hinders the model's ability to understand spatial relationships, track object transitions, or infer occlusion and continuity, particularly in long-range, egocentric sequences. To address this issue, we propose \textsc{Spatiotemporal Encoding} by explicitly encoding the motion information into the keyframes. Specifically, two trajectory-aware markers are integrated into each selected keyframe: 1) A frame index representing its position within the motion sequence.
2) A color-coded marker derived from a continuous colormap, indicating temporal progression along the 3D camera motion trajectory.
These visual markers are directly overlaid on the image, resulting in augmented frames where each carries both semantic richness (via object presence) and spatiotemporal context (via visual cues). 

In detail, given the corresponding camera trajectory and the key-frame indices $\mathcal{T}_{\text{sel}} = \{\tau_i\}_{i=0}^{K-1}$ from Section~\ref{sec:Spatial Visibility}, we assign each keyframe $\tau_i \in \mathcal{T}_{\text{sel}}$ a unique RGB color $\mathbf{c}_i$ from the same colormap as the original trajectory, i.e., $\mathbf{c}_i = \Phi(t_i) = \text{ColorMap}(\frac{t_i}{K})$. This ensures a smooth color gradient that reflects the temporal ordering and spatial progression along the motion path. Each frame $f_{\tau_i}$ is then augmented with a simple inpainting operation: we draw a filled circle at a fixed top-right position with color $\mathbf{c}_i$ and overlay the frame index $t_i$ as a label inside the marker (as shown in the bottom-right corner of Figure \ref{fig:framework}). The resulting modified frame $f'_{\tau_i}$ compactly encodes both the temporal sequence and spatial trajectory through these visual cues.

This transformation preserves the integrity of selected frames while adding motion cues, enhancing multimodal models' ability to associate frames with motion progression and spatial layout. Unlike implicit positional encoding or post-hoc temporal reasoning, our approach offers a direct, efficient, and interpretable way to inject trajectory awareness into the input space. Consequently, \textsc{MLLMs} gain a deeper understanding of scene evolution and the spatial distribution of visual content.

\vspace{-0.5em}
\subsection{Joint Optimized Prompting}
\vspace{-0.5em}
\label{sec:JointOptimizedPrompting}

We leverage the enhanced keyframes obtained from Maximum Semantic Richness Sampling (Sec.~\ref{sec:Spatial Visibility}) alongside RGB visualizations of the BEV and 3D trajectories generated from Motion Reconstruction (Sec.~\ref{sec:Motion Reconstruction}) to construct a unified visual input. These elements are then combined with a carefully designed textual prompt that includes: 1) general descriptions of the selected keyframes and their semantic properties, 2) references to the BEV and 3D trajectory visualizations, and (3) the relative coordinates of key points along the camera path. Together with a target spatial reasoning question, these components are integrated into an instruction-style prompt. This heuristic formulation serves to inject explicit spatial cues into the \textsc{MLLM}’s input, enhancing its ability to reason about geometric layout, motion progression, and scene structure in a lightweight, model-agnostic manner. Note that our method is a training- and GPU-free method, which just needs to compute several times on the CPU and requires only a single forward pass.

\begin{table}[t]
\centering
\caption{Comparison of various \textsc{MLLMs} boosting by \textbf{\textsc{See\&Trek}} on the \textsc{VSI-Bench} benchmark. $^\dag$ indicates results on \textsc{VSI-Bench} (tiny) set. ${^*}$ indicates we use the instruct version.}
\resizebox{\linewidth}{!}{
\begin{tabular}{l|c|cccccccc}
 \multirow{2}{*}{\textbf{Methods}} & \multirow{2}{*}{\textbf{Avg.}} & Obj. Count & Abs. Dist. & Obj. Size & Room Size & Rel. Dist. & Rel. Dir. & Route Plan & Appr. Order \\
 & & \multicolumn{4}{c}{\cellcolor{orange!10}\textbf{Numerical Answer}} & \multicolumn{4}{c}{\cellcolor{yellow!10}\textbf{Multiple-Choice Answer}} \\
\midrule
\rowcolor{gray!20} \multicolumn{10}{l}{\textit{Baseline}} \\
\textsc{Chance Level (Random)} & - & - & - & - & - & 25.0 & 36.1 & 28.3 & 25.0 \\
\textsc{Chance Level (Frequency)} & 34.0 & 62.1 & 32.0 & 29.9 & 33.1 & 25.1 & 47.9 & 28.4 & 25.2 \\
\midrule
\rowcolor{gray!20} \multicolumn{10}{l}{\textit{VSI-Bench (tiny) Perf.}} \\
$^\dag$\textsc{Human Level} & 79.2 & 94.3 & 47.0 & 60.4 & 45.9 & 94.7 & 95.8 & 95.8 & 100.0 \\
$^\dag$\textsc{Gemini-1.5 Flash} & 45.7 & 50.8 & 33.6 & 56.5 & 45.2 & 48.0 & 39.8 & 32.7 & 59.2 \\
$^\dag$\textsc{Gemini-1.5 Pro} & 48.8 & 49.6 & 28.8 & 58.6 & 49.4 & 46.0 & 48.1 & 42.0 & 68.0 \\
$^\dag$\textsc{Gemini-2.0 Flash} & 45.4 & 52.4 & 30.6 & 66.7 & 31.8 & 56.0 & 46.3 & 24.5 & 55.1 \\
\midrule
\rowcolor{gray!20} \multicolumn{10}{l}{\textit{Proprietary Models (API)}} \\
\textsc{Gemini-1.5 Flash} & 42.1 & 49.8 & 30.8 & 53.5 & 54.4 & 37.7 & 41.0 & 31.5 & 37.8 \\
\textsc{Gemini-1.5 Pro} & 45.4 & 56.2 & 30.9 & 64.1 & 43.6 & 51.3 & 46.3 & 36.0 & 34.6 \\
\textsc{GPT-4o} & 34.0 & 46.2 & 5.3 & 43.8 & 38.2 & 37.0 & 41.3 & 31.5 & 28.5 \\
\midrule
\rowcolor{gray!20} \multicolumn{10}{l}{\textit{Open-source Models}} \\
\textsc{LLaVA-OneVision-0.5B} & 27.1 & 33.3 & 29.2 & 13.0 & 29.1 & 29.2 & 40.2 & 36.1 & 6.6 \\
\rowcolor{blue!10} \hspace{1em}+\textbf{\textsc{See\&Trek}} & 28.7$_{\textcolor{red}{+1.6\%}}$ & 49.0$_{\textcolor{red}{+16.3\%}}$ & 29.4$_{\textcolor{red}{+0.2\%}}$ & 15.1$_{\textcolor{red}{+2.1\%}}$ & 27.7$_{\textcolor{darkgreen}{-1.4\%}}$ & 30.3$_{\textcolor{red}{+1.1\%}}$ & 37.3$_{\textcolor{darkgreen}{-2.9\%}}$ & 35.137.3$_{\textcolor{darkgreen}{-1.0\%}}$ & 6.1$_{\textcolor{darkgreen}{-0.5\%}}$ \\
\textsc{LLaVA-OneVision-7B} & 31.4 & 34.7 & 20.6 & 47.3 & 18.1 & 40.4 & 32.4 & 32.5 & 24.9
 \\
\rowcolor{blue!10} \hspace{1em}+\textbf{\textsc{See\&Trek}} & 33.0$_{\textcolor{red}{+1.6\%}}$ & 32.0$_{\textcolor{darkgreen}{-2.7\%}}$ & 17.0$_{\textcolor{darkgreen}{-3.6\%}}$ & 39.8$_{\textcolor{darkgreen}{-7.5\%}}$ & 27.8$_{\textcolor{red}{+9.7\%}}$ & 39.0$_{\textcolor{darkgreen}{-1.4\%}}$ & 40.6$_{\textcolor{red}{+8.2\%}}$ & 31.9$_{\textcolor{darkgreen}{-0.6\%}}$ & 35.7$_{\textcolor{red}{+10.8\%}}$ \\
\textsc{LLaVA-NeXT-Video-7B} & 32.5 & 37.0 & 12.1 & 45.6 & 26.4 & 35.1 & 39.3 & 34.0 & 31.1 \\
\rowcolor{blue!10} \hspace{1em}+\textbf{\textsc{See\&Trek}} & 33.8$_{\textcolor{red}{+1.3\%}}$ & 39.2$_{\textcolor{red}{+2.2\%}}$ & 11.9$_{\textcolor{darkgreen}{-0.2\%}}$ & 47.5$_{\textcolor{red}{+1.9\%}}$ & 26.6$_{\textcolor{red}{+0.2\%}}$ & 39.9$_{\textcolor{red}{+4.8\%}}$ & 40.9$_{\textcolor{red}{+1.6\%}}$ & 36.6$_{\textcolor{red}{+2.6\%}}$ & 28.3$_{\textcolor{darkgreen}{-2.8\%}}$ \\
\textsc{InternVL3-1B} & 29.5 & 65.0 & 18.5 & 15.9 & 22.5 & 29.3 & 47.8 & 27.3 & 9.8 \\
\rowcolor{blue!10} \hspace{1em}+\textbf{\textsc{See\&Trek}} & 32.0$_{\textcolor{red}{+3.5\%}}$ & 63.6$_{\textcolor{darkgreen}{-1.4\%}}$ & 25.8$_{\textcolor{red}{+7.4\%}}$ & 16.1$_{\textcolor{red}{+0.2\%}}$ & 30.7$_{\textcolor{red}{+8.2\%}}$ & 32.8$_{\textcolor{red}{+10.0\%}}$ & 46.4$_{\textcolor{darkgreen}{-1.4\%}}$ & 28.4$_{\textcolor{red}{+1.1\%}}$ & 12.3$_{\textcolor{red}{+2.5\%}}$ \\
\textsc{InternVL3-8B} & 40.2 & 67.8 & 32.2 & 44.5 & 41.8 & 42.8 & 37.7 & 26.3 & 28.3 \\
\rowcolor{blue!10} \hspace{1em}+\textbf{\textsc{See\&Trek}} & 43.2$_{\textcolor{red}{+3.0\%}}$ & 65.2$_{\textcolor{darkgreen}{-2.6\%}}$ & 32.9$_{\textcolor{red}{+0.7\%}}$ & 46.9$_{\textcolor{red}{+2.4\%}}$ & 46.7$_{\textcolor{red}{+4.9\%}}$ & 45.9$_{\textcolor{red}{+3.1\%}}$ & 40.2$_{\textcolor{red}{+2.5\%}}$ & 30.4$_{\textcolor{red}{+4.1\%}}$ & 37.4$_{\textcolor{red}{+9.1\%}}$ \\
\textsc{InternVL3-14B} & 44.2 & 69.0 & 33.6 & 53.7 & 45.6 & 43.9 & 42.4 & 23.7 & 41.4 \\
\rowcolor{blue!10} \hspace{1em}+\textbf{\textsc{See\&Trek}} & 45.6$_{\textcolor{red}{+1.4\%}}$ & 65.9$_{\textcolor{darkgreen}{-3.1\%}}$ & 35.7$_{\textcolor{red}{+2.1\%}}$ & 50.5$_{\textcolor{darkgreen}{-2.2\%}}$ & 48.4$_{\textcolor{red}{+2.8\%}}$ & 49.0$_{\textcolor{red}{+5.1\%}}$ & 41.0$_{\textcolor{darkgreen}{-1.4\%}}$ & 27.8$_{\textcolor{red}{+4.1\%}}$ & 46.8$_{\textcolor{red}{+5.4\%}}$ \\
\textsc{Qwen2.5-VL-3B} & 25.7 & 15.0 & 17.4 & 16.0 & 27.0 & 35.1 & 44.6 & 29.9 & 21.1 \\
\rowcolor{blue!10} \hspace{1em}+\textbf{\textsc{See\&Trek}} & 26.7$_{\textcolor{red}{+1.0\%}}$ & 9.7$_{\textcolor{darkgreen}{-5.3\%}}$ & 23.7$_{\textcolor{red}{+6.3\%}}$ & 19.0$_{\textcolor{red}{+3.0\%}}$ & 22.7$_{\textcolor{darkgreen}{-4.3\%}}$ & 33.2$_{\textcolor{darkgreen}{-1.9\%}}$ & 47.0$_{\textcolor{red}{+2.4\%}}$ & 29.4$_{\textcolor{darkgreen}{-0.5\%}}$ & 28.8$_{\textcolor{red}{+7.7\%}}$ \\
\textsc{Qwen2.5-VL-7B} & 27.3 & 13.0 & 14.4 & 35.9 & 21.3 & 36.9 & 37.9 & 29.9 & 29.6\\
\rowcolor{blue!10} \hspace{1em}+\textbf{\textsc{See\&Trek}} & 29.0$_{\textcolor{red}{+2.6\%}}$ & 13.6$_{\textcolor{red}{+0.6\%}}$ & 14.7$_{\textcolor{red}{+0.3\%}}$ & 35.4$_{\textcolor{darkgreen}{-0.5\%}}$ & 23.6$_{\textcolor{red}{+2.3\%}}$ & 33.4$_{\textcolor{darkgreen}{-3.5\%}}$ & 41.3$_{\textcolor{red}{+3.4\%}}$ & 30.4$_{\textcolor{red}{+0.5\%}}$ & 39.2$_{\textcolor{red}{+9.6\%}}$ \\
\textsc{Qwen2.5-VL-32B} & 33.7 & 16.7 & 22.6 & 47.0 & 37.8 & 39.3 & 37.6 & 29.9 & 38.7 \\
\rowcolor{blue!10} \hspace{1em}+\textbf{\textsc{See\&Trek}} & 34.7$_{\textcolor{red}{+1.0\%}}$ & 19.9$_{\textcolor{red}{+3.2\%}}$ & 23.8$_{\textcolor{red}{+1.2\%}}$ & 41.0$_{\textcolor{darkgreen}{-6.0\%}}$ & 39.5$_{\textcolor{red}{+1.7\%}}$ & 36.9$_{\textcolor{darkgreen}{-2.4\%}}$ & 39.2$_{\textcolor{red}{+1.6\%}}$ & 32.5$_{\textcolor{red}{+2.6\%}}$ & 44.9$_{\textcolor{red}{+6.2\%}}$ \\
\textsc{${^*}$Kimi-VL-A3B} & 33.4 & 24.2 & 30.8 & 49.6 & 33.5 & 33.8 & 39.7 & 28.9 & 27.2 \\
\rowcolor{blue!10} \rowcolor{blue!10} \hspace{1em}+\textbf{\textsc{See\&Trek}} & 35.1$_{\textcolor{red}{+1.7\%}}$ & 23.5$_{\textcolor{darkgreen}{-0.7\%}}$ & 30.4$_{\textcolor{darkgreen}{-0.4\%}}$ & 48.4$_{\textcolor{darkgreen}{-1.2\%}}$ & 38.3$_{\textcolor{red}{+4.8\%}}$ & 35.6$_{\textcolor{red}{+1.8\%}}$ & 41.9$_{\textcolor{red}{+2.2\%}}$ & 30.9$_{\textcolor{red}{+2.0\%}}$ & 31.9$_{\textcolor{red}{+4.7\%}}$ \\
\bottomrule
\end{tabular}}
\label{tab:vsibench}
\vspace{-1.5em}
\end{table}

\vspace{-0.75em}
\section{Experiments}
\vspace{-0.5em}

\textit{
Implementation and more comprehensive experiments details can be found in Appendix~\ref{app:experiments}.}

\vspace{-0.75em}
\subsection{Evaluation Setup}
\vspace{-0.5em}

\textbf{Datasets.} We select \textsc{VSI-Bench}~\cite{vsibench} and \textsc{STI-Bench}~\cite{STIBench} as our spatial evaluation benchmark. 
\textbf{1) \textsc{VSI-Bench}} is a very challenging benchmark that requires understanding spatial relationships and correspondences of multiple objects in a video~\cite{vsibench}. It comprises over 5,000 question-answer pairs derived from 288 real videos, with duration ranging from 1.5 minutes to 2.5 minutes. These videos are sourced from the validation sets of the public indoor 3D scene reconstruction datasets ScanNet~\cite{Scannet} , ScanNet++~\cite{yeshwanth2023scannet++}, and ARKitScenes~\cite{Arkitscenes} and represent diverse environments—including residential spaces, professional settings (\textit{e.g.}, offices, labs), and industrial spaces (e.g., factories)—and multiple geographic regions. Compared to normal multimodal spatial benchmarks like \textsc{ScanQA}~\cite{Scanqa}, \textsc{VSI-Bench} has a wider range of random changes in viewing angles. \textbf{2) \textsc{STI-Bench}} is a benchmark designed to evaluate MLLMs’ spatial-temporal understanding through challenging tasks such as estimating and predicting the appearance, pose, displacement, and motion of objects. It contains 2,064 QA pairs across desktop, indoor, and outdoor scenarios, providing a systematic quantitative assessment of MLLMs’ spatial-temporal understanding capabilities. 

\textbf{Benchmark Models.} Following~\cite{vsibench}, we comprehensively evaluate 10 video-supporting open-source MLLMs across diverse model families on our proposed \textbf{\textsc{See\&Trek}}, encompassing various parameter scales, training recipes, and model architectures. For proprietary models, we consider Gemini-1.5~\cite{team2024gemini} and GPT-4o~\cite{gpt} for comparison. For open-source models, we evaluate models from InternVL3~\cite{zhu2025internvl3}, LLaVA-OneVision~\cite{Llava-onevision}, LLaVA-NeXT-Video~\cite{LLaVA-NeXT}, Qwen2.5~\cite{qwen25} and Kimi-VL~\cite{kimi}. All evaluations are conducted under zero-shot settings. To ensure reproducibility, we use greedy decoding for all models. In \textsc{VSI-Bench} evaluation, \textit{Baseline} and \textit{VSI-Bench (tiny) Perf.} are borrowed from~\cite{vsibench}, which are only utilized for comparison. 

\begin{table}[t]
\centering
\caption{Comparison of various \textsc{MLLMs} boosting by \textbf{\textsc{See\&Trek}} on the \textsc{STI-Bench}.}
\resizebox{\linewidth}{!}{
\begin{tabular}{l|c|cccccccc}

 \multirow{2}{*}{\textbf{Methods}} & \multirow{2}{*}{\textbf{Avg.}}  & \makecell{Dim.\\Meas.} & \makecell{Spatial\\Rel.} & \makecell{3D Video\\Grounding} & \makecell{Disp.\\\& P.L.} & \makecell{Speed\\\& Acc.} & \makecell{Ego\\\& Orient.} & \makecell{Traj.\\Desc.} & \makecell{Pose\\Est.} \\

 & & \multicolumn{3}{c}{\cellcolor{orange!10}\textbf{Static Understanding}} & \multicolumn{5}{c}{\cellcolor{yellow!10}\textbf{Dynamic Understanding}} \\
\midrule
\rowcolor{gray!20} \multicolumn{10}{l}{\textit{Proprietary Models (API)}} \\
GPT-4o & 34.8 & 24.9 & 49.6 & 28.1 & 27.6 & 36.0 & 30.3 & 36.8 & 51.3 \\
Claude-3.7-Sonnet & 39.4 & 31.8 & 49.0 & 36.3 & 29.0 & 36.9 & 27.0 & 41.0 & 62.7 \\
Gemini-2.0-Flash & 38.7 & 33.7 & 50.0 & 33.7 & 32.7 & 34.4 & 15.1 & 48.7 & 62.4 \\
Gemini-2.5-Pro & 40.9 & 34.2 & 53.4 & 32.3 & 32.4 & 34.3 & 44.9 & 52.0 & 58.4 \\
\midrule
\rowcolor{gray!20} \multicolumn{10}{l}{\textit{Open-source Models}} \\
\textsc{InternVL3-1B} & 18.7 & 19.4 & 19.2 & 18.6 & 19.3 & 18.8 & 8.1 & 24.4 & 21.7 \\
\rowcolor{blue!10} \hspace{1em}+\textbf{\textsc{See\&Trek}} & 20.4$_{\textcolor{red}{+1.7\%}}$ & 25.0$_{\textcolor{red}{+5.6\%}}$ & 17.4$_{\textcolor{darkgreen}{-1.8\%}}$ & 20.5$_{\textcolor{red}{+1.9\%}}$ & 18.2$_{\textcolor{darkgreen}{-1.1\%}}$ & 19.7$_{\textcolor{red}{+0.9\%}}$ & 12.6$_{\textcolor{red}{+4.5\%}}$ & 22.4$_{\textcolor{darkgreen}{-2.0\%}}$ & 24.4$_{\textcolor{red}{+2.7\%}}$ \\
\textsc{InternVL3-8B} & 30.2 & 27.0 & 36.3 & 28.7 & 25.2 & 32.7 & 23.8 & 25.6 & 42.2 \\
\rowcolor{blue!10} \hspace{1em}+\textbf{\textsc{See\&Trek}} & 31.2$_{\textcolor{red}{+1.0\%}}$ & 26.7$_{\textcolor{darkgreen}{-0.3\%}}$ & 36.4$_{\textcolor{red}{+0.1\%}}$ & 34.1$_{\textcolor{red}{+5.4\%}}$ & 21.0$_{\textcolor{darkgreen}{-3.8\%}}$ & 35.4$_{\textcolor{red}{+2.7\%}}$ & 27.9$_{\textcolor{red}{+4.1\%}}$ & 26.3$_{\textcolor{red}{+0.7\%}}$ & 42.7$_{\textcolor{red}{+0.5\%}}$ \\
\textsc{InternVL3-14B} & 30.8 & 27.7 & 41.1 & 32.2 & 19.1 & 28.8 & 17.3 & 25.6 & 49.4 \\
\rowcolor{blue!10} \hspace{1em}+\textbf{\textsc{See\&Trek}} & 32.2$_{\textcolor{red}{+1.4\%}}$ & 29.9$_{\textcolor{red}{+2.2\%}}$ & 40.3$_{\textcolor{darkgreen}{-0.8\%}}$ & 32.2$_{\textcolor{red}{+0.0\%}}$ & 22.4$_{\textcolor{red}{+3.3\%}}$ & 30.5$_{\textcolor{red}{+1.7\%}}$ & 19.7$_{\textcolor{red}{+2.4\%}}$ & 34.2$_{\textcolor{red}{+8.6\%}}$ & 48.0$_{\textcolor{darkgreen}{-1.4\%}}$ \\
\textsc{Qwen2.5-VL-7B} & 35.6 & 25.3 & 52.1 & 33.4 & 19.9 & 31.5 & 41.6 & 50.0 & 52.2
 \\
\rowcolor{blue!10} \hspace{1em}+\textbf{\textsc{See\&Trek}} & 36.9$_{\textcolor{red}{+1.3\%}}$ & 24.2$_{\textcolor{darkgreen}{-1.1\%}}$ & 45.2$_{\textcolor{darkgreen}{-6.9\%}}$ & 35.3$_{\textcolor{red}{+1.9\%}}$ & 20.5$_{\textcolor{red}{+0.6\%}}$ & 32.1$_{\textcolor{red}{+0.6\%}}$ & 57.8$_{\textcolor{red}{+16.2\%}}$ & 48.7$_{\textcolor{darkgreen}{-1.3\%}}$ & 52.2$_{\textcolor{black}{+0.0\%}}$ \\
\textsc{Qwen2.5-VL-32B} & 40.5 & 36.3 & 46.6 & 39.7 & 33.6 & 40.0 & 22.7 & 44.8 & 57.2
 \\
\rowcolor{blue!10} \hspace{1em}+\textbf{\textsc{See\&Trek}} & 41.7$_{\textcolor{red}{+1.2\%}}$ & 36.7$_{\textcolor{red}{+0.4\%}}$ & 49.3$_{\textcolor{red}{+2.7\%}}$ & 37.2$_{\textcolor{darkgreen}{-2.5\%}}$ & 31.7$_{\textcolor{darkgreen}{-1.9\%}}$ & 42.4$_{\textcolor{red}{+2.4\%}}$ & 33.5$_{\textcolor{red}{+10.8\%}}$ & 46.2$_{\textcolor{red}{+1.4\%}}$ & 58.9$_{\textcolor{red}{+1.7\%}}$ \\
\bottomrule
\end{tabular}}
\label{tab:STI-Bench}
\vspace{-0.4cm}
\end{table}

\vspace{-0.5em}
\subsection{Main Results}
\vspace{-0.5em}

\textbf{1) \textsc{VSI-Bench}.} As shown in Table \ref{tab:vsibench}, \textbf{\textsc{See\&Trek}} consistently enhances the performance of various open-source multimodal models on the \textsc{VSI-Bench} benchmark. In terms of overall accuracy (Avg.), all tested models benefit from the integration of \textbf{\textsc{See\&Trek}}, with gains ranging from +1.0\% to +3.5\%. The most notable improvement is observed on \textsc{InternVL3-1B}, which achieves a +3.5\% boost, highlighting the effectiveness of \textbf{\textsc{See\&Trek}} for relatively smaller models. Regarding specific task types,  \textbf{\textsc{See\&Trek}} notably improves numerical reasoning (\textit{e.g.}, Abs. Dist.) and spatial understanding (\textit{e.g.}, Route Plan and Appr. Order), where we observe significant relative gains such as +10.8\% in Appr. Order for \textsc{LLaVA-OneVision-7B}. These trends suggest that \textbf{\textsc{See\&Trek}} is particularly beneficial for complex spatial-temporal reasoning tasks. We also observe that larger models (\textit{e.g.}, \textsc{InternVL3-14B}, \textsc{Qwen2.5-VL-32B}) tend to show modest but consistent improvements. This indicates that while \textbf{\textsc{See\&Trek}} helps models of all sizes, its relative impact is more pronounced on lightweight or mid-sized MLLMs, potentially due to their higher reliance on external structural cues to compensate for limited capacity. It is also worth noting that some performance drops (highlighted in green) appear in isolated tasks (\textit{e.g.}, Rel. Dir. and Obj. Count) for certain models, which might be attributed to trade-offs introduced by the augmented perception pipeline or model-specific biases. However, the overall trend strongly favors the inclusion of \textbf{\textsc{See\&Trek}} as a plug-and-play enhancement module for open-source MLLMs in video-centric spatial reasoning tasks.

\textbf{2) \textsc{STI-Bench}.} As shown in Table \ref{tab:STI-Bench}, the integration of \textbf{\textsc{See\&Trek}} consistently enhances overall accuracy across all evaluated open-source models on the \textsc{STI-Bench}. For the \textsc{InternVL3} series, they achieve respective improvements of +1.7\%, +1.0\%, and +1.4\% in average performance respectively. This consistent gain across models of varying scales underscores the scalability and robustness of \textbf{\textsc{See\&Trek}}. At the sub-task level, \textbf{\textsc{See\&Trek}} yields particularly notable improvements in dynamic understanding. For instance, \textsc{InternVL3-14B} shows substantial gains in Trajectory Description (+8.6\%) and Displacement \& Path Length (+3.3\%), reflecting enhanced temporal and spatial tracking capabilities. Likewise, \textsc{InternVL3-8B} benefits significantly in 3D Video Grounding (+5.4\%) and Ego \& Orientation (+2.7\%). These results demonstrate that \textbf{\textsc{See\&Trek}} is especially effective at reinforcing temporal-spatial reasoning. Although minor performance declines are observed in a few categories (\textit{e.g.}, Room Size or Relative Direction), they are marginal and do not offset the substantial improvements in key dynamic tasks.

\begin{table}[ht]
\centering
\caption{Ablation studies of Maximum Semantic Richness Sampling and Motion Reconstruction.}
\vspace{-0.2cm}
\resizebox{\linewidth}{!}{
\begin{tabular}{cc|c|cccccccc}
    \toprule
 Max. Semantic & Motion & \multirow{2}{*}{\textbf{Avg.}}  & Obj. Count & Abs. Dist. & Obj. Size & Room Size & Rel. Dist. & Rel. Dir. & Route Plan & Appr. Order \\
 Rich. Sampling & Reconstruction  & & \multicolumn{4}{c}{\cellcolor{orange!10}\textbf{Numerical Answer}} & \multicolumn{4}{c}{\cellcolor{yellow!10}\textbf{Multiple-Choice Answer}} \\
 \midrule
   &  & 40.2 & 67.8 & 32.2 & 44.5 & 41.8 & 42.8 & 37.7 & 26.3 & 28.3 \\
  \checkmark & & 41.8 & 67.3 & 34.7 & 39.0  & 47.5 & 43.9 & 44.3 & 25.8 & 31.9 \\
   & \checkmark & 42.1 & 64.6 & 29.7 & 46.4 & 44.7 & 39.1 & 40.7 & 28.8 & 43.0 \\ 
  \checkmark & \checkmark & 43.2 & 65.2 & 32.9 & 46.9 & 46.7 & 45.9 & 40.2 & 30.4 & 37.4 \\
\bottomrule
\end{tabular}}
\label{tab:abl_overall}
\end{table}

\vspace{-0.25em}
\subsection{Ablation Studies} 
\vspace{-0.25em}

In this section, we utilize \textsc{VSI-Bench} for evaluation and leverage \textsc{InternVL3-8B} as the baseline model. We investigate the effect of each technique proposed in our \textbf{\textsc{See\&Trek}}. \textit{More investigation can be found in the Appendix.}

\textbf{Overall.} 
To systematically assess the contribution of different spatial cues, we first conduct ablation studies on two components: Maximum Semantic Richness Sampling (dubbed MSRS) and Motion Reconstruction. As shown in Table \ref{tab:abl_overall}, enabling MSRS alone leads to a moderate performance gain (average accuracy from 40.2\% to 41.8\%), with notable improvements observed in static layout-related tasks such as object size. This suggests that semantically richer keyframe selection contributes to more informative spatial representations. Besides, incorporating Motion Reconstruction yields distinct advantages in dynamic reasoning tasks, particularly relative direction and route planning (\textit{e.g.}, route planning improves from 26.3\% to 28.8\%, highlighting its role in modeling egocentric movement and temporal coherence. When both components are jointly applied, the model achieves the highest overall accuracy (43.2\%), demonstrating their complementary effects in facilitating both static and dynamic aspects of spatial understanding within \textsc{MLLMs}.

\textbf{Optimized Prompting.} 
Then, we investigate the impact of \textsc{Spatiotemporal Encoding} and point prompts added in instruction from Section \ref{sec:JointOptimizedPrompting}.
As shown in Table \ref{tab:abl_remain}, adding \textsc{Spatiotemporal Encoding} improves the average accuracy to 42.7\%, with notable gains in tasks like Object Count (66.3\%) and Relative Direction (46.9\%), highlighting its role in enhancing spatial and temporal reasoning which connecting the MSRS and Motion Reconstruction. Similarly, incorporating point prompts also achieves an average accuracy of 42.7\% demonstrating its effectiveness in providing explicit spatial cues. When both modules are combined, the model achieves the highest average accuracy of 43.2\%, with significant improvements in diverse tasks. 

\begin{table}[t]
\centering
\caption{Ablation studies of \textsc{Spatiotemporal Encoding} and point prompts.}
\resizebox{\linewidth}{!}{
\begin{tabular}{cc|c|cccccccc}
\toprule
 \textsc{Spatiotemporal} & point & \multirow{2}{*}{\textbf{Avg.}}  & Obj. Count & Abs. Dist. & Obj. Size & Room Size & Rel. Dist. & Rel. Dir. & Route Plan & Appr. Order \\
 \textsc{Encoding} & prompts  & & \multicolumn{4}{c}{\cellcolor{orange!10}\textbf{Numerical Answer}} & \multicolumn{4}{c}{\cellcolor{yellow!10}\textbf{Multiple-Choice Answer}} \\
 \midrule
   &  & 42.4 & 64.9 & 32.8 & 45.3  & 46.9 & 44.9 & 40.5  & 27.3 & 37.1 \\
  \checkmark &  &  42.7 & 66.3 & 34.7 & 43.5 & 46.9 & 46.9 & 39.1 & 31.4 & 34.7 \\
   & \checkmark & 42.7 & 66.5 & 35.0 & 44.1 & 44.9 & 46.6 & 38.9 & 29.9 & 35.4 \\
  \checkmark & \checkmark & 43.2 & 65.2 & 32.9 & 46.9 & 46.7 & 45.9 & 40.2 & 30.4 & 37.4 \\
\bottomrule
\end{tabular}}
\label{tab:abl_remain}
\vspace{-0.2cm}
\end{table}

\textbf{Sample Efficiency Analysis.} We conduct experiments with different sample intervals $N \in (1, 2, 3, 4, 8, 12)$  from a single forward process to analyze their impact on the efficiency performance of \textbf{\textsc{See\&Trek}}. While the finest granularity ($N=1$) achieves strong results (Avg. 42.9\%) at the cost of high computational time (410s), increasing $N$ substantially reduces runtime while maintaining competitive accuracy. $N=3$ and $N=4$ strike the best balance, achieving the highest average score (43.2\%) with over 65\% reduction in processing time compared to $N=1$. Even with larger intervals like $N=8$ or $N=12$, the model retains robust spatial understanding, with only a marginal drop in performance. These results demonstrate that \textbf{\textsc{See\&Trek}} is highly sample-efficient, maintaining strong accuracy with significantly fewer frames and minimal computational overhead. \textit{Note that the time consuming of \textbf{\textsc{See\&Trek}} still largely depends the length of the given videos. Particularly, the duration of videos in \textsc{VSI-Bench} at least longer than 1 minute.}

\begin{table}[ht]
\centering
\caption{Ablation studies of different sample interval $N$ settings. ``Time(s)'' denotes the average time consuming on processing videos from \textsc{VSI-Bench}.}
\resizebox{\linewidth}{!}{
\begin{tabular}{cc|c|cccccccc}
\toprule
\multirow{2}{*}{$N$} & \multirow{2}{*}{Time(s)} & \multirow{2}{*}{\textbf{Avg.}} & Obj. Count & Abs. Dist. & Obj. Size & Room Size & Rel. Dist. & Rel. Dir. & Route Plan & Appr. Order \\
 &  & & \multicolumn{4}{c}{\cellcolor{orange!10}\textbf{Numerical Answer}} & \multicolumn{4}{c}{\cellcolor{yellow!10}\textbf{Multiple-Choice Answer}} \\
 \midrule
    1 & 410 & 42.9 & 66.2 & 31.7 & 48.2 & 47.3 & 42.8 & 42.0 & 30.4 & 33.8 \\
    2 & 227 & 41.7 & 66.0 & 30.5 & 47.1 & 46.3 & 40.0 & 40.8 & 27.8 & 35.3 \\
    3 & 159 & 43.2 & 66.9 & 31.3 & 47.3 & 48.4 & 43.8 & 40.6 & 29.4 & 38.3 \\
    4 & 82 & 43.2 & 65.2 & 32.9 & 46.9 & 46.7 & 45.9 & 40.3 & 30.4 & 37.4 \\
    8 & 63 & 42.8 & 66.5 & 33.6 & 46.0 & 46.7 & 41.7 & 41.5 & 30.9 & 35.9 \\
    12 & 53 & 42.5 & 66.4 & 32.6 & 45.9 & 41.7 & 46.5 & 39.9 & 28.9 & 38.8 \\
\bottomrule
\end{tabular}}
\label{tab:sample}
\vspace{-0.4cm}
\end{table}

\begin{table}[ht]
\centering
\caption{Ablation studies of different rank extraction methods in MSRS.}
\resizebox{\linewidth}{!}{
\begin{tabular}{c|c|cccccccc}
    \toprule
Rank & \multirow{2}{*}{\textbf{Avg.}} & Obj. Count & Abs. Dist. & Obj. Size & Room Size & Rel. Dist. & Rel. Dir. & Route Plan & Appr. Order \\
 Extraction  &  & \multicolumn{4}{c}{\cellcolor{orange!10}\textbf{Numerical Answer}} & \multicolumn{4}{c}{\cellcolor{yellow!10}\textbf{Multiple-Choice Answer}} \\
 \midrule
\textit{Top$K$} & 37.9 & 65.9 & 28.7 & 42.6 & 35.5 & 37.1 & 39.2 & 28.9 & 25.7 \\
\textit{Temporal-Top$K$} & 42.7 & 69.0 & 31.5 & 47.0 & 45.5 & 41.0 & 39.7 & 31.4 & 36.9 \\
\textit{Balanced-Top$K$} & 43.2 & 65.2 & 32.9 & 46.9 & 46.7 & 45.9 & 40.2 & 30.4 & 37.4 \\
\bottomrule
\end{tabular}}
\label{tab:rank}
\vspace{-0.2cm}
\end{table}

\textbf{Rank Selection.} We explore different rank extraction methods' impact on MSRS. Here, we utilize other two kinds of Top$K$ selection methods: 1) Original \textit{Top$K$} only extracts the frames according to most object numbers. 2) \textit{Temporal-Top$K$} temporally divides a video into multiple consecutive frame groups based on the number of keyframes and then performs \textit{Top$K$} selection within each group. As shown in Table \ref{tab:rank}, \textit{Balanced-Top$K$} consistently outperforms both \textit{Top$K$} and \textit{Temporal-Top$K$} across various spatial reasoning tasks, demonstrating its advantage in capturing both comprehensive and diverse visual semantics. Notably, it achieves the highest average accuracy (43.2\%) and excels in multiple-choice tasks such as Relative Distance (45.9\%) and Approach Order (37.4\%), indicating a better spatial understanding through a more balanced frame selection strategy. Compared with the \textit{Top$K$} and \textit{Temporal-Top$K$}, our proposed \textit{Balanced-Top$K$} further enhances selection by jointly considering object richness, temporal distribution, and semantic diversity.

\vspace{-0.25em}
\subsection{Qualitative results}
\vspace{-0.25em}

\begin{figure}[t]
    \centering
    \includegraphics[width=\linewidth]{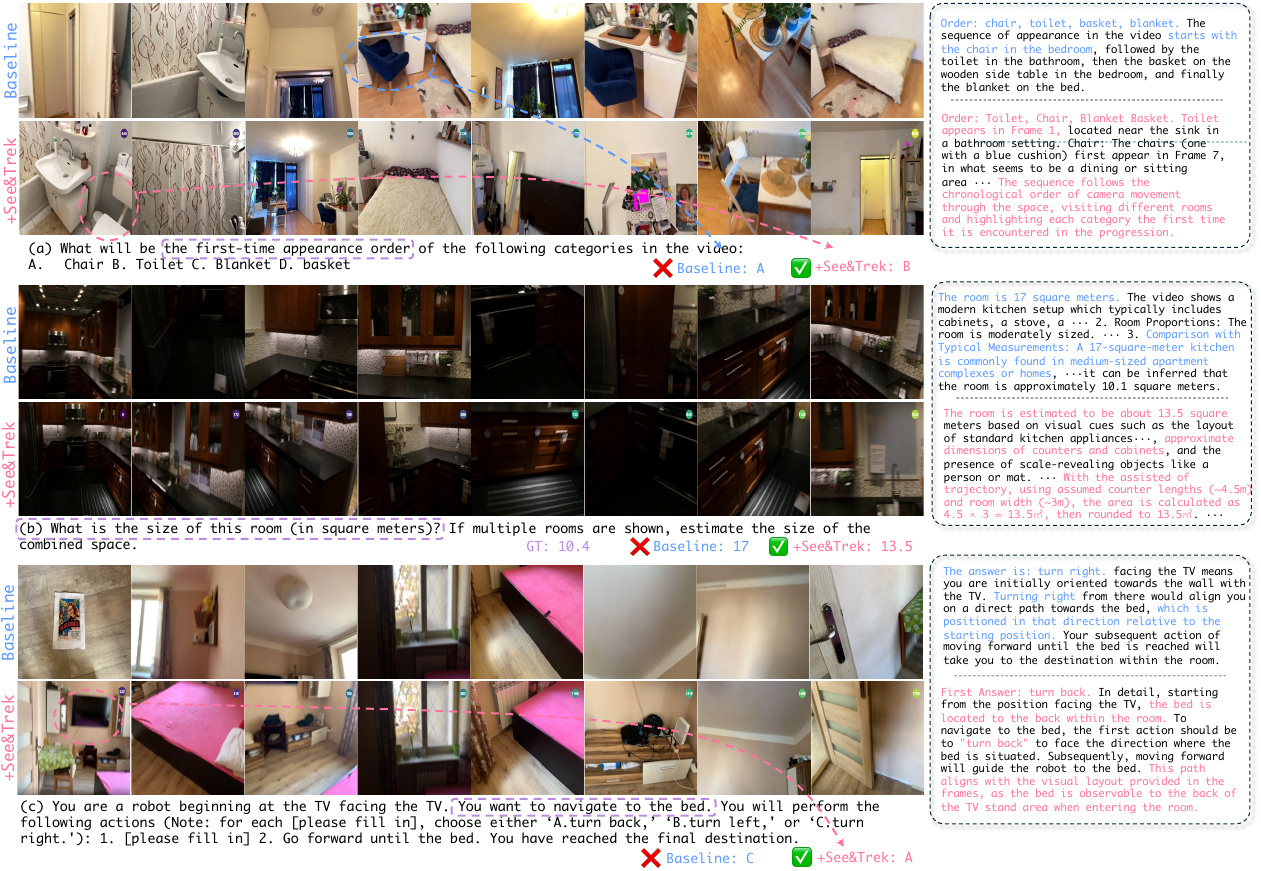}
    \vspace{-0.6cm}
    \caption{Qualitative results about \textbf{\textsc{See\&Trek}} evaluated on \textsc{VSI-Bench}. Here, we represent the different tasks of (a) appearance order (b) room size, and (c) route plan. \textit{More results can be found in the appendix.} It shows that \textbf{\textsc{See\&Trek}} obtains the visual diversity and motion reconstruction of the given video, gaining a better spatial understanding.}
    \label{fig:results}
    \vspace{-0.5em} 
\end{figure}

As shown in Figure \ref{fig:results}, we leverage self-explanations to investigate why \textbf{\textsc{See\&Trek}} achieves superior performance. Thanks to the MSRS, \textbf{\textsc{See\&Trek}} is able to identify the most informative frame that best represents the entire scene—such as in cases (a) and (c)—typically from the first frame, while the baseline always fails in \textcolor{paperblue}{perception and positioning}. This leads to more accurate analysis in dynamic spatial understanding tasks for \textsc{MLLMs}. Furthermore, we observe that motion reconstruction enhances the MLLM's ability to estimate spatial distances with lower error, as demonstrated in case (b), like \textcolor{paperred}{``with the assisted of trajectory,''} thereby supporting static spatial reasoning. Overall, \textbf{\textsc{See\&Trek}} significantly enhances the comprehensive spatial understanding capabilities of \textsc{MLLMs}.

\vspace{-0.75em}
\section{Conclusion}
\vspace{-0.5em}
We propose \textbf{\textsc{See\&Trek}}, the first training- \& GPU-free prompting framework designed to enhance the spatial understanding capabilities of \textsc{MLLMs}. We focus on enhancing the spatial understanding ability of MLLMs from two aspects: Visual Diversity and Motion Reconstruction. For achieving Visual Diversity, we conduct Maximum Semantic Richness Sampling, which employs an off-the-shell perception model, \textit{e.g.} \textsc{YOLO} to efficiently extract frames that maximize semantic richness, then propose \textit{Balanced-Top$K$} strategy for selecting frames. For Motion Reconstruction, we simulate visual trajectories and encode relative spatial positions to preserve both spatial relations and temporal coherence. Our method is GPU-free, requiring only a single forward pass, and can be seamlessly integrated into existing \textsc{MLLMs}. Comprehensive experiments on various \textsc{MLLMs} and two hard spatial benchmarks verify the \textbf{\textsc{See\&Trek}} superiority in boosting spatial intelligence. 

\clearpage

\newpage
\appendix

\section{Methods}
\subsection{Detailed Algorithm}
The detailed algorithm of \textbf{\textsc{See\&Trek}} could be found as follows:
\begin{algorithm}[ht] \small
\label{alg:detailed}
\caption{\textsc{See\&Trek}:  Maximum Semantic Richness Sampling and Motion Reconstruction}
\label{alg:seeandtrek}
\begin{algorithmic}[1]
\Require Video $V = \{f_t\}_{t=1}^{N_v}$, sampling interval $N$, number of keyframes $K$, object detector $\mathcal{Y}(\cdot)$, camera intrinsics $\mathbf{K}$
\Ensure Selected keyframes $\{\mathcal{F}'_{t_i}\}_{i=1}^{K}$, BEV trajectory $\mathbf{P}_{\text{BEV}}$, 3D trajectory $\mathbf{P}_{\text{3D}}$, 3D trajectory set $\mathcal{T}^{\text{world}}$  
\State Sample frames: $\mathcal{S} \gets \{f_t \mid t \bmod N = 0\}$
\State Initialize: $\mathcal{O} \gets \emptyset$, Trajectory $\mathcal{T}^{\text{world}} \gets []$
\For{each consecutive pair $(f_{t-1}, f_t) \in \mathcal{S}$}
    \State Convert $(f_{t-1}, f_t)$ to grayscale
    \State Extract ORB keypoints/descriptors: $\text{ORB}(f_{t-1}), \text{ORB}(f_t)$
    \State Match descriptors to obtain $\mathcal{M}_{t-1,t}$
    \State Normalize matched points: $\hat{\mathbf{x}}_i \gets \mathbf{K}^{-1}[\mathbf{x}_i^\top\; 1]^\top$
    \State Estimate essential matrix $\mathbf{E}$ via RANSAC
    \State Decompose $\mathbf{E} \rightarrow (\mathbf{R}_t, \mathbf{T}_t)$
    \State Update global pose: $\mathbf{R}_t^{\text{world}}, \mathbf{T}_t^{\text{world}}$
    \State Append $\mathbf{T}_t^{\text{world}}$ to $\mathcal{T}^{\text{world}}$

    \State Detect objects: $\mathcal{C}_t \gets \mathcal{Y}(f_t)$
    \State Append $(t, \mathcal{C}_t)$ to $\mathcal{O}$
\EndFor
\State Filter valid interval $[t_s, t_e]$: $\mathcal{O}^* \gets \{(t_i, \mathcal{C}_{t_i}) \in \mathcal{O} \mid t_s \leq t_i \leq t_e\}$
    \State \textbf{Balanced-Top$K$ Selection}:
    \State Select global-rich frame: $t_g \gets \arg\max |\mathcal{C}_{t_i}|$
    \State $\mathcal{C}_{\text{sel}} \gets \mathcal{C}_{t_g}$, $\mathcal{T}_{\text{sel}} \gets \{t_g\}$
    \State Divide $\mathcal{O}^*$ into $K{-}1$ segments $\{\mathcal{O}^{(1)},\dots,\mathcal{O}^{(K-1)}\}$
    \For{$k = 1$ to $K-1$}
        \State $t_k \gets \arg\min_{(t_j, \mathcal{C}_{t_j}) \in \mathcal{O}^{(k)}} \left(|\mathcal{C}_{t_j} \cap \mathcal{C}_{\text{sel}}|,\ -|\mathcal{C}_{t_j}|,\ t_j\right)$
        \State $\mathcal{C}_{\text{sel}} \gets \mathcal{C}_{\text{sel}} \cup \mathcal{C}_{t_k}$, $\mathcal{T}_{\text{sel}} \gets \mathcal{T}_{\text{sel}} \cup \{t_k\}$
    \EndFor
\State \textbf{Spatiotemporal Encoding:}
\For{each $t_i \in \mathcal{T}_{\text{sel}}$}
    \State Compute color $\mathbf{c}_i \gets \text{ColorMap}(t_i/K)$
    \State Overlay index and color on $f_{t_i} \Rightarrow \mathcal{F}'_{t_i}$
\EndFor
\State \textbf{Trajectory Visualization:} $\mathbf{P}_{\text{BEV}} \gets \text{ProjectXY}(\mathcal{T}^{\text{world}})$, $\mathbf{P}_{\text{3D}} \gets \text{Render3D}(\mathcal{T}^{\text{world}})$
\end{algorithmic}
\end{algorithm}

\subsection{Motion Reconstruction}

In this section, we recall the mathematical context of Visual Odometry (VO). \textit{Note that most of these can be found in the textbooks.} The first step of VO is need to conduct ORB (Oriented FAST and Rotated BRIEF) Feature Detection and Description, then Feature Matching, finally perform Essential Matrix Estimation. For this part of algorithm development, we leverage OpenCV \footnote{\url{https://github.com/opencv/opencv-python}} for efficient deployment. 

\textbf{ORB Feature Detection and Description.} In detailed, ORB combines FAST keypoint detection with orientation-augmented BRIEF descriptors, producing a scale- and rotation-invariant feature. Regarding \textbf{Keypoint Detection (FAST)}: Given a grayscale image $f_t: \Omega \subset \mathbb{R}^2 \rightarrow \mathbb{R}$, the FAST corner detector selects a pixel $\mathbf{p} \in \Omega$ as a keypoint if there exists a contiguous arc of $n$ pixels on the Bresenham circle $C(\mathbf{p})$ of radius 3 around $\mathbf{p}$, such that:
\begin{equation}
    |f_t(\mathbf{q})-f_t(\mathbf{p})|>\tau,\quad\forall\mathbf{q}\in C_n(\mathbf{p}),
\end{equation}
where $C_n(\mathbf{p}) \subset C(\mathbf{p})$ is a contiguous segment of $n$ pixels, and $\tau$ is a contrast threshold. Typically, $n=12$ and $|C(\mathbf{p})| = 16$. This step yields the raw keypoint set $\mathcal{K}_t=\left\{\mathbf{p}_i^t\in\Omega\right\}_{i=1}^N,$ where $N$ is the number of detected corners in frame $f_t$.
Regarding \textbf{Orientation Assignment}, to achieve rotation invariance, ORB computes the orientation angle $\theta_i$ of each keypoint $\mathbf{p}_i^t$ by using intensity moments of a patch $P_i$ around the keypoint:
\begin{equation}
    \theta_i=\arctan\left(\frac{m_{01}}{m_{10}}\right),\quad m_{pq}=\sum_{\mathbf{q}\in P_i}x^py^qf_t(\mathbf{q}),
\end{equation}
where $(x, y)$ are coordinates relative to the keypoint $\mathbf{p}_i^t$. Finally, we perform \textbf{Descriptor Computation (BRIEF)}. The BRIEF descriptor $\mathcal{D}_t \in {0,1}^{N \times D}$ is constructed by binary intensity comparisons between $D/2$ pre-defined point pairs $(\mathbf{a}_k, \mathbf{b}_k)$ in the local patch around each $\mathbf{p}_i^t$, rotated by angle $\theta_i$: 
\begin{equation}
\mathcal{D}_t(i,k)=\begin{cases}1&\mathrm{if~}f_t(R_{\theta_i}\mathbf{a}_k+\mathbf{p}_i^t)<f_t(R_{\theta_i}\mathbf{b}_k+\mathbf{p}_i^t),\\0&\mathrm{otherwise},\end{cases}
\end{equation}
where $R_{\theta_i} \in SO(2)$ is the 2D rotation matrix corresponding to $\theta_i$.

\textbf{Feature Matching.} Given two sets of ORB descriptors $\mathcal{D}_{t-1}, \mathcal{D}_t \in {0,1}^{N \times D}$, feature correspondences are established by computing Hamming distances between binary descriptor vectors. For each descriptor $\mathbf{d}i^{t-1} \in \mathcal{D}{t-1}$, define the matching descriptor in frame $t$ as:
\begin{equation}
    \mathbf{d}_j^t=\arg\min_{\mathbf{d}\in\mathcal{D}_t}\mathrm{Hamming}(\mathbf{d}_i^{t-1},\mathbf{d}),
\end{equation}
and accept the match if the distance is below a threshold $\tau_d$ or passes Lowe's ratio test. Let the resulting matched keypoint pairs be:
\begin{equation}
    \mathcal{M}_{t-1,t}=\left\{(\mathbf{x}_i^{t-1},\mathbf{x}_i^t)\right\}_{i=1}^P,\quad\mathbf{x}_i^{t-1},\mathbf{x}_i^t\in\mathbb{R}^2,
\end{equation}
with $\mathbf{x}i^{t-1} = \mathbf{p}{a_i}^{t-1}$ and $\mathbf{x}i^t = \mathbf{p}{b_i}^{t}$ corresponding to matched keypoints.

\textbf{Essential Matrix Estimation.} Given matched pixel coordinates ${(\mathbf{x}_i^{t-1}, \mathbf{x}i^t)}{i=1}^P$ and the intrinsic matrix $\mathbf{K} \in \mathbb{R}^{3 \times 3}$, normalized coordinates are computed: 
\begin{equation}
    \hat{\mathbf{x}}_i^{t-1}=\mathbf{K}^{-1}\begin{bmatrix}\mathbf{x}_i^{t-1}\\1\end{bmatrix},\quad\hat{\mathbf{x}}_i^t=\mathbf{K}^{-1}\begin{bmatrix}\mathbf{x}_i^t\\1\end{bmatrix},\quad\hat{\mathbf{x}}\in\mathbb{R}^3.
\end{equation}
The essential matrix $\mathbf{E} \in \mathbb{R}^{3 \times 3}$ encodes the relative motion such that $(\hat{\mathbf{x}}_i^t)^\top\mathbf{E}\hat{\mathbf{x}}_i^{t-1}=0,\forall i.$ In matrix terms, stacking all equations gives:
\begin{equation}
    \mathbf{A}\cdot\mathrm{vec}(\mathbf{E})=0,\mathrm{where~}\mathbf{A}\in\mathbb{R}^{P\times9}.
\end{equation}
This is solved via SVD of $\mathbf{A}$ in the 8-point algorithm, or a robust estimator such as:
\begin{equation}
\mathbf{E}=\arg\min_{\mathbf{E}^{\prime}}\sum_{i}\rho\left((\hat{\mathbf{x}}_{i}^{t})^{\top}\mathbf{E^{\prime}}\hat{\mathbf{x}}_{i}^{t-1}\right),
\end{equation}
where $\rho(\cdot)$ is a robust loss (e.g., truncated quadratic), and outliers are rejected via RANSAC. Since $\mathbf{E}$ must satisfy the singular value constraint $\sigma_1 = \sigma_2, \sigma_3 = 0$, we project it onto the essential matrix manifold via SVD:
\begin{equation}
    \mathbf{E}=\mathbf{U}\cdot\mathrm{diag}(1,1,0)\cdot\mathbf{V}^{\top}.
\end{equation}
Finally, we decompose the essential matrix to motion. Specially, The essential matrix relates to the relative pose via: $\mathbf{E}=[\mathbf{T}_{t}]_{\times}\mathbf{R}_{t},$ where $[\cdot]_\times$ denotes the skew-symmetric matrix: 
\begin{equation}
    [\mathbf{T}_{t}]_{\times}=\begin{bmatrix}0&-T_{z}&T_{y}\\T_{z}&0&-T_{x}\\-T_{y}&T_{x}&0\end{bmatrix}.
\end{equation}
The decomposition of $\mathbf{E}$ via SVD provides four candidate solutions $(\pm \mathbf{R}_t, \pm \mathbf{T}_t)$, disambiguated by the Cheirality condition—checking the number of triangulated points in front of both cameras.

\subsection{Joint Optimized Prompting}

\begin{figure}[t]
    \centering
    \includegraphics[width=\linewidth]{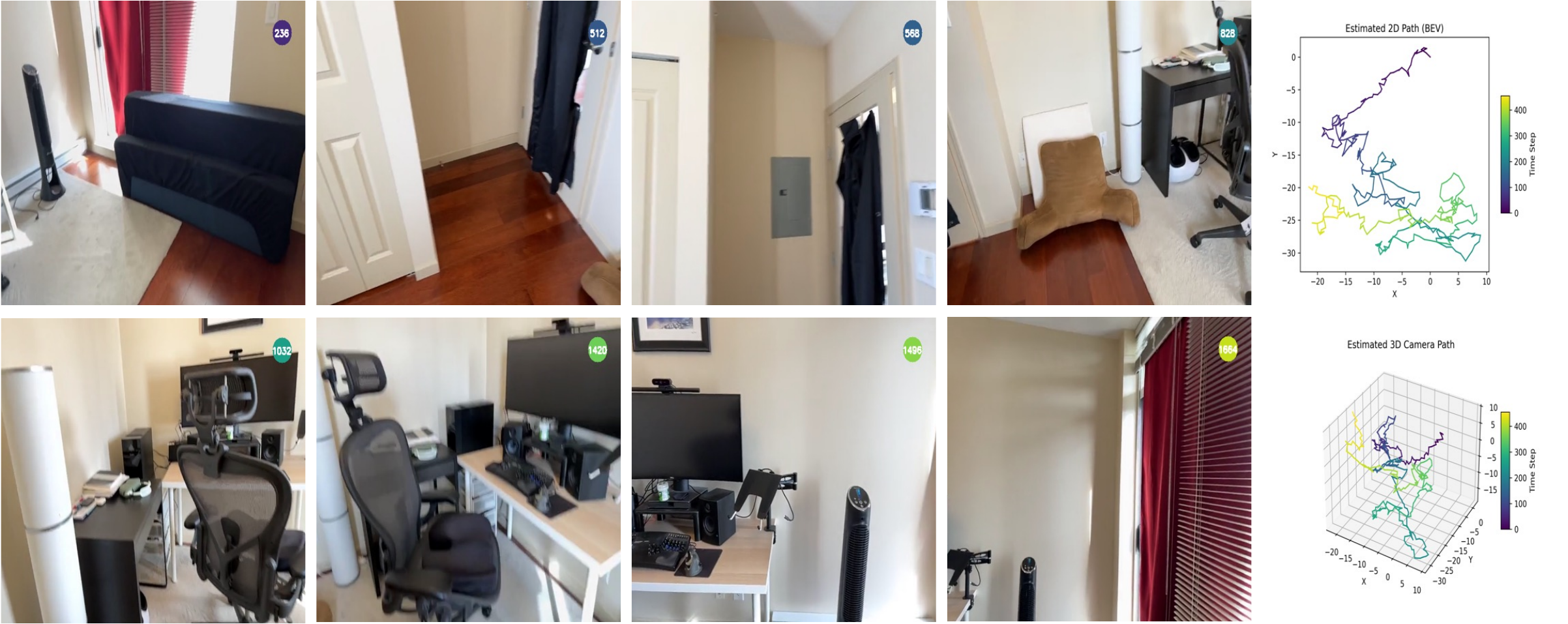}
    \vspace{-0.4cm}
    \caption{The sample of the visual prompting input in \textbf{\textsc{See\&Trek}}.}
    \label{fig:visual-prompt}
\end{figure}

Here, we give the sample of our general input containing text and vision for \textsc{MLLMs} evaluation. For \textsc{Internvl3-8B}, after spatial prompting from \textbf{\textsc{See\&Trek}}, the visual inputs are as shown in Figure \ref{fig:visual-prompt}, then we also design the corresponding text prompt template as shown in \ref{lst:textprompt}. ``Points'' denotes the relative spatial coordinates of each selected frames. Since different MLLMs have different training data, architectures, and training methods, using text prompts or point coordinates as prompts has a particularly large impact on performance. Therefore, we fine-tune the instructions for different model series to enable \textbf{\textsc{See\&Trek}} to fully utilize its spatial understanding advantages.

\begin{figure}[ht]
\centering
\begin{lstlisting}[language=Python, label={lst:textprompt}]
spatial_prompt_universal = (
    "Each video frame has its serial number in the top-right corner. "
    "The highlight color mark of frame matches the color in the spatial map, indicating its position."
)
spatial_prompt_2D_3D = (
    "Both 2D (bird's-eye) and 3D views illustrate the camera's spatial trajectory, "
    "with color encoding time progression."
)
spatial_prompt_points = (
    "Points represent the camera's relative positions; the number of points reflects only spatial relationships."
)
input_prompt = spatial_prompt_universal + spatial_prompt_2D_3D + spatial_prompt_points + points + "\n" + question 
\end{lstlisting}
\caption{The sample of the text prompting input in \textbf{\textsc{See\&Trek}}. ``Points'' denotes the relative spatial coordinates of each selected frames.}
\end{figure}

\section{Experiments}
\label{app:experiments}

\subsection{VSI-Bench Dataset}

\textbf{Overview.} \textsc{VSI-Bench}~\cite{vsibench} includes eight tasks of three types: \textit{configurational}, \textit{measurement estimation}, and \textit{spatiotemporal}. The configurational tasks (\textit{object count}, \textit{relative distance}, \textit{relative direction}, \textit{route plan}) test a model’s understanding of the configuration of space and are more intuitive for humans. Measurement estimation (of \textit{object size}, \textit{room size}, and \textit{absolute distance}) is of value to any embodied agent. While predicting a measurement exactly is very difficult, for both humans and models, a better sense of distance and other measurements is intuitively correlated with better visual-spatial intelligence and underpins a wide range of tasks that require spatial awareness, like interaction with objects and navigation. Spatiotemporal tasks like appearance order test a model’s memory of space as seen in the video. 

\textbf{Metric Design.} Based on whether the ground-truth answer is verbal or numerical, \textsc{VSI-Bench} tasks are suited to either a Multiple-Choice Answer (MCA) or Numerical Answer (NA) format. For MCA tasks, \textsc{VSI-Bench} follows standard practice by using \textit{Accuracy} ($\mathcal{ACC}$), based on exact matching (with possible fuzzy matching), as the primary metric. For NA tasks, where models predict continuous values, accuracy via exact matching fails to capture the degree of proximity between model predictions and ground-truth answers. Therefore,  \textsc{VSI-Bench} utilizes a new metric, \textit{Mean Relative Accuracy} ($\mathcal{MRA}$) inspired by previous works. Specifically, for a NA question, given a model’s prediction $\hat{y}$, ground truth $y$, and a confidence threshold $\theta$, relative accuracy is calculated by considering $\hat{y}$ correct if the relative error rate, defined as $|\hat{y}-y|/y$, is less than $1-\theta$. As single-confidence-threshold accuracy only considers relative error in a narrow scope, $\mathcal{MRA}$ averages the relative accuracy across a range of confidence thresholds $\mathcal{C}=\{0.5,0.55,...,0.95\}$:
\begin{equation}
    \mathcal{MRA}=\frac{1}{10}\sum_{\theta\in\mathcal{C}}\mathbb{I}\left(\frac{|\hat{y}-y|}{y}<1-\theta\right).
\end{equation}
$\mathcal{MRA}$ offers a more reliable and discriminative measurement for calculating the similarity between numerical predictions and ground truth values.

\textbf{Chance Level Baselines}. \textsc{VSI-Bench} provides two baselines: 1) \textit{Chance Level} (Random) is the random selection accuracy for MCA tasks (and is inapplicable for NA tasks). 2) \textit{Chance Level} (Frequency) represents the highest performance \textsc{MLLMs} would achieve by always selecting the most frequent answer for each task. This identifies performance gains that may result from inherently long-tailed answers or imbalanced multiple-choice distributions.

\textbf{Human Level Performance.} \textsc{VSI-Bench} randomly sample a subset of 400 questions (50 per task), which we will refer to as \textsc{VSI-Bench} (tiny). Human evaluators independently answer each question, and their performance is evaluated using the above-mentioned metrics.

\begin{figure}[t]
    \centering
    \includegraphics[width=\linewidth]{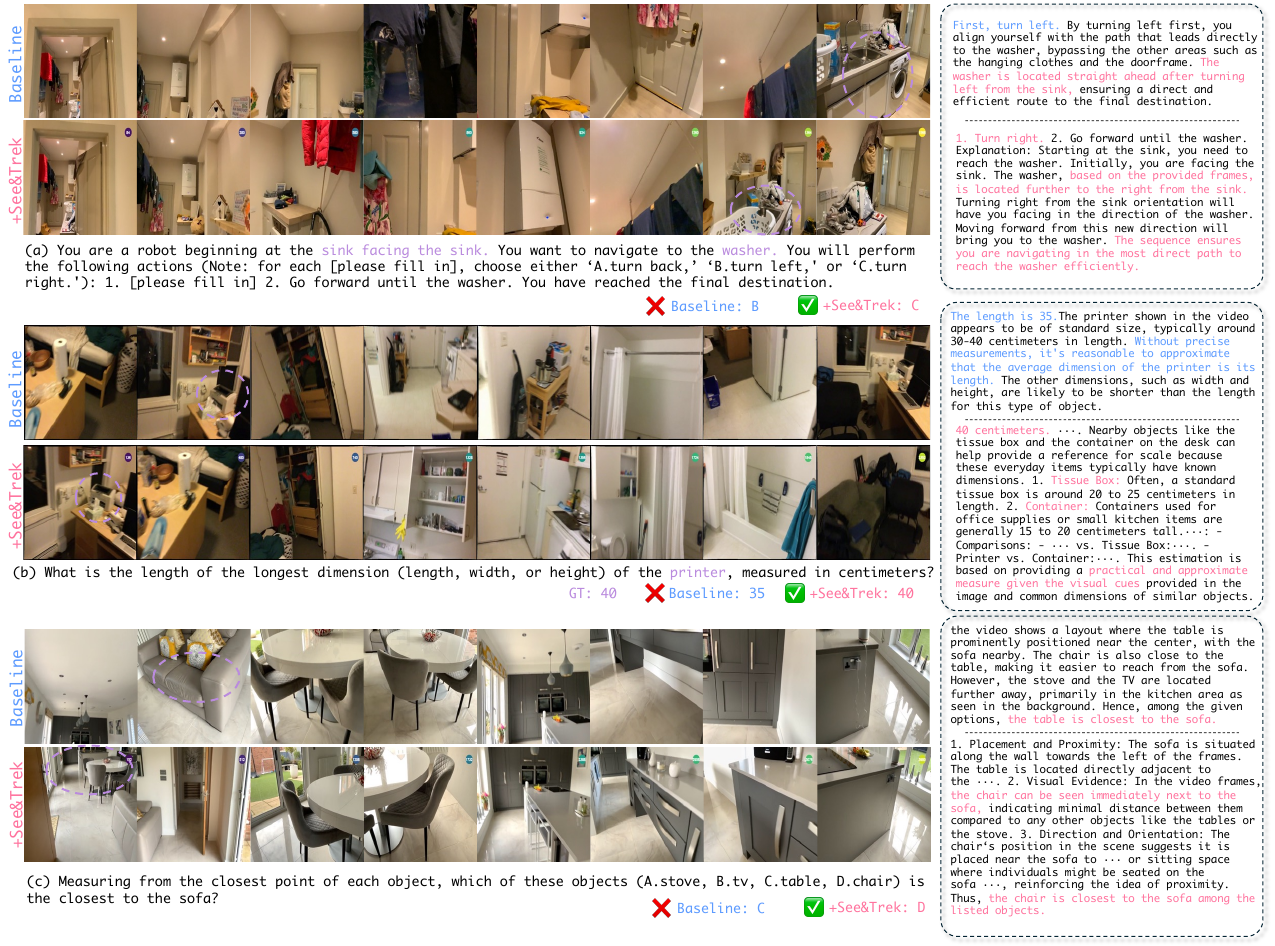}
    \vspace{-0.4cm}
    \caption{Qualitative results about \textbf{\textsc{See\&Trek}} evaluated on \textsc{VSI-Bench}. Here, we represent the different tasks of (a) Route Plan (b) Object Absolute Size (Obj. Size), and (c) Object Relative Distance (Rel. Dist.).}
    \label{fig:results_1}
\end{figure}

\begin{figure}[t]
    \centering
    \includegraphics[width=\linewidth]{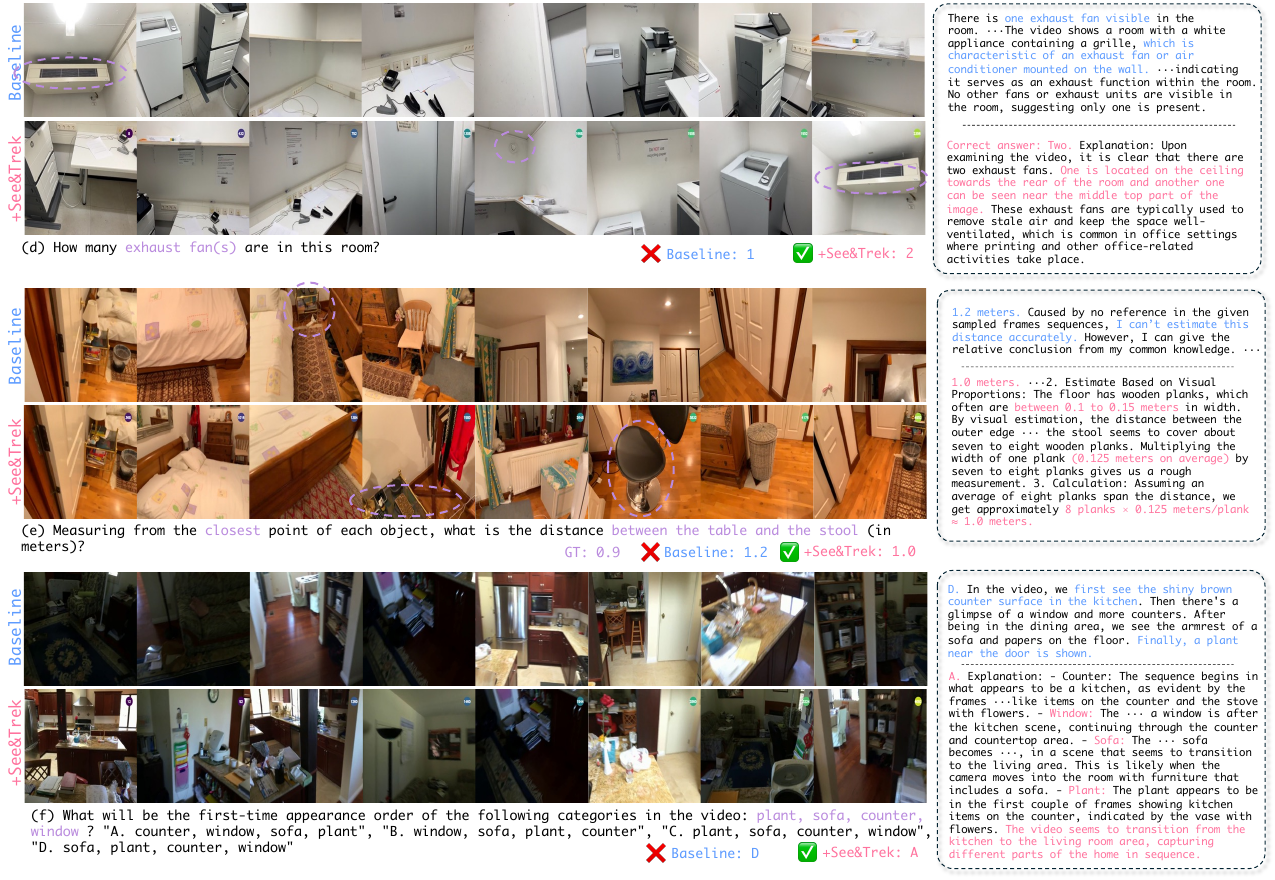}
    \vspace{-0.4cm}
    \caption{Qualitative results about \textbf{\textsc{See\&Trek}} evaluated on \textsc{VSI-Bench}. Here, we represent the different tasks of (d) Object Count (Obj. Count) (e) Object Absolute Distance (Abs. Dist.), and (f) Appearance Order (Appr. Order).}
    \label{fig:results_2}
\end{figure}

\begin{figure}[t]
    \centering
    \includegraphics[width=\linewidth]{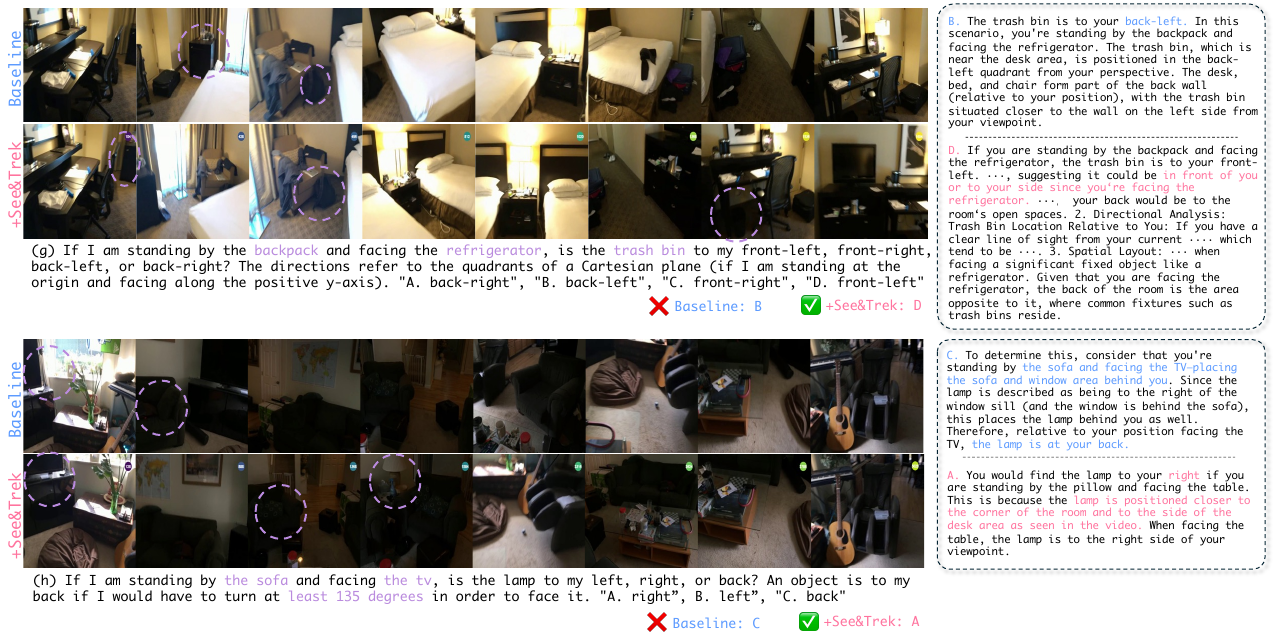}
    \vspace{-0.4cm}
    \caption{Qualitative results about \textbf{\textsc{See\&Trek}} evaluated on \textsc{VSI-Bench}. Here, we represent the different tasks of (g) Route Plan (h) Object Relative Direction (Rel. Dir.)}
    \label{fig:results_3}
\end{figure}

\subsection{STI-Bench Dataset}

\textbf{Overview.} \textsc{STI-Bench} \cite{STIBench} contains 300 videos and more than 2,000 QA pairs, covering three major scenarios: Desktop, Indoor, and Outdoor. The videos are sourced from \textsc{Omni6DPose} \cite{Omni6dpose}, \textsc{ScanNet} \cite{Scannet} and \textsc{Waymo} \cite{Waymo} respectively, thus encompassing a broad spectrum of real-world environments. They propose eight tasks in total, each one systematically examining a distinct aspect of MLLMs’ spatial-temporal understanding, which divide these tasks into two main categories: 1) \textbf{Static Understanding}: \textit{Dimensional Measurement, Spatial Relation, 3D Video Grounding}; 2) \textbf{Dynamic Understanding}: \textit{Displacement and Path Length}, \textit{Speed and Acceleration, Ego-Centric Orientation, Trajectory Description} and \textit{Pose Estimation.} 

\textbf{API Engine and Baseline Settings.} Regarding the commercial engine, \textsc{STI-Bench} uniformly sample 30 frames from the video for each record and explicitly indicate the sampling FPS (Frames Per Second) for the current video within the prompt. An exception is made for Claude3.7-Sonnet, for which only 20 frames are sampled due to its API constraints. \textsc{STI-Bench} are presented in a multiple-choice format with five possible answers, hence a random guess baseline yields a 20\% accuracy.

\textbf{Metric Design.} 1) \textit{Dimensional Measurement.} let $l_x, l_y, l_z$ denote the dimensions (length, width, height) of an object along the $x,y,$ and $z$ axes:
\begin{equation}
    l_{x}=x_{\max}-x_{\min},l_{y}=y_{\max}-y_{\min},l_{z}=z_{\max}-z_{\min},
\end{equation}
Here, $l_x, l_y, l_z$ represent the object size along each coordinate axis. If it needs the distance between two objects (or between the camera and an object), let $d_{12}$ be the Euclidean distance between their center points:
\begin{equation}
    d_{12}=\sqrt{(x_2-x_1)^2+(y_2-y_1)^2+(z_2-z_1)^2}.
\end{equation}
Here, $(x_1, y_1, z_1)$ and $(x_2, y_2, z_2)$ are the center coordinates of the two objects.

2) \textit{Spatial Relation.} When the difference along one coordinate axis is significantly larger than along others, the sign of that difference determines the spatial relation:
\begin{equation}
    r_{xy}=\mathrm{sign}(x_A-x_B),r_{yz}=\mathrm{sign}(y_A-y_B),r_{zx}=\mathrm{sign}(z_A-z_B).
\end{equation}
Here, $r_{xy},r_{yz},r_{zx}$ indicate relative positioning along each axis (e.g., front/back, left/right, above/below). It choose  the axis with the greatest difference to label the dominant relation. 

3) \textit{3D Video Grounding.} For frame $t$ in the camera coordinate system, the 3D bounding box of an object can be described with dimensions, center position, and optional rotations: 
\begin{equation}
    \mathrm{BBox}_t=(l_t,w_t,h_t,x_t,y_t,z_t,\mathrm{yaw}_t,\mathrm{pitch}_t,\mathrm{roll}_t).
\end{equation}
Here, $(l_t,w_t,h_t)$ are the object dimensions, $(x_t,y_t,z_t)$ is the center position, and (yaw$_{t}$, pitch$_{t}$, roll$_{t}$ ) are optional rotation angles if available.

4) \textit{Pose Estimation.} Given the camera’s initial pose $(p_0,o_0)$, the pose $(p_t,o_t)$ at time $t$ can be obtained using the extrinsic-derived matrices $R_t$ (rotation) and $T_t$ (translation):
\begin{equation}
    p_t=R_tp_0+T_t,o_t =o_0+\Delta o_t.
\end{equation}
Here, $p_t$ is the position, $o_t$ is the orientation.

5) \textit{Displacement and Path Length.} Let $p_i=(x_i,y_i,z_i)$ be the position at time $i$. The displacement $d_{0n}$ and path length $L_{\mathrm{traj}}$ are computed as: 
\begin{equation}
    d_{0n}=\sqrt{(x_n-x_0)^2+(y_n-y_0)^2+(z_n-z_0)^2},
\end{equation}
\begin{equation}
    L_{\mathrm{traj}}=\sum_{i=1}^n\sqrt{(x_i-x_{i-1})^2+(y_i-y_{i-1})^2+(z_i-z_{i-1})^2}.
\end{equation}
Here, $d_{0n}$ is the straight-line distance from the initial to the final position; $L_{\mathrm{traj}}$ sums consecutive segment lengths for the entire path. 

6) \textit{Speed and Acceleration.} Let $\Delta t$ be the time interval between consecutive frames. Then the speed $v_i$ and acceleration $a_i$ are:
\begin{equation}
    v_i=\frac{d_i}{\Delta t}, a_i=\frac{v_i-v_{i-1}}{\Delta t}.
\end{equation}
Here, $d_i$ is the displacement between adjacent frames, $v_i$ is the speed at time $i$, and $a_i$ is the acceleration.

7) \textit{Ego-Centric Orientation.} If $\theta_t$ denotes the camera orientation (azimuth) at time $t$, then the orientation change $\Delta \theta_t$ is:
\begin{equation}
    \Delta\theta_t=\theta_t-\theta_0.
\end{equation}
This indicates how much the camera has rotated relative to its initial azimuth.

8) \textit{Trajectory Description.} It applies the Ramer-DouglasPeucker (RDP) algorithm to simplify the sequence of positions into key line segments. The resulting polyline is described in a piecewise manner (\textit{e.g.}, ``go straight for 30m, turn left 85°, then go straight for 20m, ...''), providing a concise representation of complex motion trajectories.

\subsection{Implementation}
\textbf{\textsc{See\&Trek}} focus on the pre-processing stage of \textsc{MLLMs}, which samples one frame for every four frames from the given spatial video. For fair evaluation, we adopt 8 frames as input to test each \textsc{MLLMs} for the given videos. For obtaining visual diversity and balancing efficiency, we utilize \textsc{YOLOv8}-Tiny, named \textsc{YOLOv8n} from Ultralytics\footnote{\url{https://github.com/ultralytics/ultralytics}} for faster detection. All experiments are conducted on NVIDIA 8$\times$A6000 and 6$\times$A800. In actual development, to accelerate the overall testing process, we first utilize Motion construction and YOLO to process each sampled frame from videos for subsequent evaluation calls, which stores the corresponding spatial information. In evaluation, we call the stored spatial information from the last step and perform the \textit{Balanced-TopK} sampling and \textsc{Spatiotemporal Encoding} techniques in each question and answer process.

\subsection{Ablation Study}

In this section, we utilize \textsc{VSI-Bench} for evaluation and leverage \textsc{InternVL3-8B} as the baseline model. We investigate the effect of each technique proposed in our \textbf{\textsc{See\&Trek}}.

\begin{table}[ht]
\centering
\caption{Ablation studies of each component of \textit{Balanced-Top$K$}.}
\resizebox{\linewidth}{!}{
\begin{tabular}{c|c|cccccccc}
    \toprule
\multirow{2}{*}{\textit{Balanced-Top$K$}} & \multirow{2}{*}{\textbf{Avg.}} & Obj. Count & Abs. Dist. & Obj. Size & Room Size & Rel. Dist. & Rel. Dir. & Route Plan & Appr. Order \\
  &  & \multicolumn{4}{c}{\cellcolor{orange!10}\textbf{Numerical Answer}} & \multicolumn{4}{c}{\cellcolor{yellow!10}\textbf{Multiple-Choice Answer}} \\
 \midrule
\textit{w/o $\mathcal{C}_{\text{sel}}$ Min.} & 42.5 & 69.4 & 33.5 & 45.6 & 44.6 & 41.4 & 39.9 & 28.4 & 37.4 \\
\textit{w/o Count Max.} & 43.1 & 67.1 & 31.9 & 45.9 & 47.5 & 42.5 & 39.5 & 29.4 & 40.9 \\
\textit{w/o Time Early} & 43.2 & 66.2 & 34.0 & 45.5 & 48.0 & 42.1 & 41.0 & 31.4 & 37.3 \\
\bottomrule
\end{tabular}}
\label{tab:TopK}
\vspace{-0.2cm}
\end{table}

\textbf{Maximum Semantic Richness Sampling.} As mentioned in Section \ref{sec:Spatial Visibility}, we want to select one frame $\tau_k$ such that the overlap of its detected categories with $\mathcal{C}_{\text{sel}}$ is minimized, classes count is maximized, and the frame is as early as possible in \textit{Balanced-Top$K$} sampling. Here, we explore each part effect on these components. Table~\ref{tab:TopK} reports the ablation results of the three components in our proposed \textit{Balanced-Top$K$} strategy: minimizing category overlap with the selected pool ($\mathcal{C}_{\text{sel}}$ Min.), maximizing the number of detected classes (Count Max.), and selecting the earliest frame in case of ties (Time Early). The complete version of our method achieves the best average performance across all question types, validating the necessity of each design component. Removing the category overlap minimization (\textit{w/o $\mathcal{C}_{\text{sel}}$ Min.}) leads to the most significant performance drop, especially in multiple-choice tasks such as "Route Plan" and "Approach Order", indicating its crucial role in encouraging semantic diversity and avoiding redundant content across selected frames. Omitting the class count maximization (\textit{w/o Count Max.}) also degrades performance, particularly in object-centric questions like "Obj. Count" and "Rel. Dist.", demonstrating that favoring frames with more detected objects helps maximize information gain per frame. Removing the temporal prioritization (\textit{w/o Time Early}) slightly affects overall performance, with minor impacts across all metrics. This suggests that while encouraging earlier frame selection helps improve temporal coherence and avoids delayed keyframe concentration, it is relatively less critical than the other two components.

\begin{table}[ht]
\centering
\caption{Ablation studies of each visual trajectories of Motion Reconstruction.}
\resizebox{\linewidth}{!}{
\begin{tabular}{c|c|cccccccc}
    \toprule
\multirow{2}{*}{Visual Trajectories} & \multirow{2}{*}{\textbf{Avg.}} & Obj. Count & Abs. Dist. & Obj. Size & Room Size & Rel. Dist. & Rel. Dir. & Route Plan & Appr. Order \\
  &  & \multicolumn{4}{c}{\cellcolor{orange!10}\textbf{Numerical Answer}} & \multicolumn{4}{c}{\cellcolor{yellow!10}\textbf{Multiple-Choice Answer}} \\
 \midrule
\textit{w/o $\mathbf{P}_{\text{BEV}}$} & 43.1 & 65.4 & 33.1 & 48.3 & 45.4 & 45.9 & 41.0 & 29.9 & 36.4 \\
\textit{w/o $\mathbf{P}_{\text{3D}}$} & 42.8 & 65.4 & 33.3 & 47.9 & 44.9 & 45.9 & 40.7 & 27.8 & 36.4 \\
\bottomrule
\end{tabular}}
\label{tab:visualtraj}
\end{table}

\textbf{Detector Choices.} Based on the ablation studies presented in Table \ref{tab:det}, it is evident that selecting the optimal detector and sampling interval $N$ requires balancing detection accuracy and computational efficiency. Among the tested YOLO variants, \textsc{YOLOv8n} consistently achieves competitive average scores across numerical and multiple-choice question categories while maintaining the lowest inference time, especially at $N=4$, where it balances a relatively high average accuracy of 43.2\% with only 82 seconds processing time per video. Although larger models like \textsc{YOLOv8s} and \textsc{YOLOv11s} occasionally reach similar or slightly higher accuracy in some metrics, their time costs are significantly higher—often an order of magnitude more—making them less practical for real-time or resource-constrained scenarios. Notably, increasing the sampling interval $N$ generally reduces computation time but can cause minor fluctuations in accuracy, with $N=4$ emerging as an optimal trade-off that avoids excessive temporal sparsity while reducing inference overhead. Moreover, while \textsc{YOLOv11n} shows marginal improvements over \textsc{YOLOv8n} in some tasks, its time cost nearly doubles. Therefore, our final choice of \textsc{YOLOv8n} with $N=4$ reflects a well-justified compromise: it offers sufficient detection performance across diverse evaluation criteria without sacrificing speed, enabling efficient processing in practical deployment. This reinforces the insight that bigger and more complex models are not always better, and careful tuning of sampling intervals combined with lightweight detectors can yield a more balanced, efficient system.

\begin{table}[ht]
\centering
\caption{Ablation studies of different sample interval $N$ settings and detector choices. ``Time(s)'' denotes the average time consuming on processing videos from \textsc{VSI-Bench}. ``N'' denotes the tiny version and ``S'' denotes the small version.}
\resizebox{\linewidth}{!}{
\begin{tabular}{ccc|c|cccccccc}
\toprule
\multirow{2}{*}{Detector} & \multirow{2}{*}{$N$} & \multirow{2}{*}{Time(s)} & \multirow{2}{*}{\textbf{Avg.}} & Obj. Count & Abs. Dist. & Obj. Size & Room Size & Rel. Dist. & Rel. Dir. & Route Plan & Appr. Order \\
 & & & & \multicolumn{4}{c}{\cellcolor{orange!10}\textbf{Numerical Answer}} & \multicolumn{4}{c}{\cellcolor{yellow!10}\textbf{Multiple-Choice Answer}} \\
 \midrule
  \multirow{6}{*}{\textsc{YOLOv8n}} & 1 & 410 & 42.9 & 66.2 & 31.7 & 48.2 & 47.3 & 42.8 & 42.0 & 30.4 & 33.8 \\
                                   & 2 & 227 & 43.1 & 65.2 & 31.6 & 46.8 & 47.5 & 41.5 & 40.7 & 30.9 & 40.8 \\
                                    & 3 & 159 & 42.3 & 66.4 & 32.1 & 46.6 & 48.1 & 40.8 & 39.8 & 30.9 & 34.0 \\
                                    & 4 & 82 & 43.2 & 65.2 & 32.9 & 46.9 & 46.7 & 45.9 & 40.3 & 30.4 & 37.4 \\
                                   & 8 & 63 & 42.8 & 66.5 & 33.6 & 46.0 & 46.7 & 41.7 & 41.5 & 30.9 & 35.9 \\
                                    & 12 & 53 & 42.5 & 66.4 & 32.6 & 45.9 & 41.7 & 46.5 & 39.9 & 28.9 & 38.8 \\
    \midrule
  \multirow{6}{*}{\textsc{YOLOv8s}} & 1 & 1989 & 42.4 & 66.2 & 31.6 & 46.9 & 46.4 & 39.4 & 41.8 & 29.9 & 37.2 \\
                               & 2 & 997 & 41.7 & 66.0 & 30.5 & 47.1 & 46.3 & 40.0 & 40.8 & 27.8 & 35.3 \\
                                & 3 & 518 & 43.2 & 66.9 & 31.3 & 47.3 & 48.4 & 43.8 & 40.6 & 29.4 & 38.3 \\
                                & 4 & 414 & 43.1 & 66.5 & 31.8 & 46.1 & 48.1 & 41.5 & 39.1 & 29.4 & 42.1 \\
                               & 8 & 236 & 41.9 & 64.6 & 29.2 & 46.1 & 43.9 & 37.2 & 41.2 & 30.4 & 43.2 \\
                                & 12 & 132 & 41.5 & 65.7 & 32.6 & 46.3 & 43.6 & 39.6 & 40.0 & 28.9 & 35.3 \\
    \midrule
  \multirow{6}{*}{\textsc{YOLOv11n}} & 1 & 397 & 42.2 & 66.4 & 32.6 & 47.1 & 45.8 & 41.0 & 41.0 & 29.9 & 34.0 \\
                               & 2 & 219 & 42.7 & 66.6 & 32.1 & 46.4 & 46.8 & 40.1 & 39.2 & 31.9 & 39.0 \\
                                & 3 & 144 & 42.6 & 66.5 & 31.2 & 46.9 & 49.2 & 41.3 & 40.3 & 30.4 & 35.4 \\
                                & 4 & 75 & 43.2 & 66.2 & 34.0 & 45.5 & 48.0 & 42.1 & 41.0 & 31.4 & 37.4 \\
                               & 8 & 58 & 42.4 & 66.9 & 31.4 & 38.7 & 45.8 & 45.8 & 33.0 & 40.1 & 38.5 \\
                                & 12 & 47 & 42.5 & 66.4 & 32.6 & 46.5 & 41.7 & 46.5 & 28.9 & 39.9 & 38.8 \\
    \midrule
  \multirow{6}{*}{\textsc{YOLOv11s}} & 1 & 1763 & 43.4 & 67.4 & 32.6 & 43.2 & 48.1 & 48.7 & 30.9 & 48.1 & 37.5 \\
                               & 2 & 940 & 43.0 & 67.1 & 32.3 & 43.5 & 47.0 & 47.2 & 28.9 & 40.8 & 37.2 \\
                                & 3 & 554 & 42.8 & 67.8 & 32.5 & 42.4 & 46.4 & 48.1 & 28.9 & 42.4 & 34.5 \\
                                & 4 & 397 & 42.4 & 66.8 & 32.5 & 41.5 & 46.4 & 48.9 & 30.9 & 39.7 & 32.5 \\
                               & 8 & 254 & 42.2 & 66.7 & 33.0 & 41.8 & 46.6 & 47.8 & 29.4 & 40.7 & 31.9 \\
                                & 12 & 113 & 41.8 & 66.5 & 33.2 & 40.7 & 46.2 & 45.2 & 28.4 & 40.4 & 34.1 \\
\bottomrule
\end{tabular}}
\label{tab:det}
\vspace{-0.2cm}
\end{table}

\textbf{Keyframe Number Analysis.} We further investigate the impact of varying the number of keyframes $\mathcal{K}$ on model performance. As shown in Table~\ref{tab:keyframe}, \textsc{See\&Trek} consistently improves spatial understanding across all keyframe settings compared to the baseline \textsc{InternVL3-8B}. The most notable gain of +3.0\% is observed at $\mathcal{K}=8$, demonstrating that our method is highly effective even under sparse temporal input. Interestingly, as $\mathcal{K}$ increases, the performance improvement becomes marginal, with the gain reduced to +0.6\% at $\mathcal{K}=32$. This diminishing return can be attributed to the saturation of spatial information in densely sampled frames: when adjacent keyframes are visually redundant, they provide limited additional cues for the model to reason over spatial relationships. In contrast, our semantic richness sampling strategy is designed to select frames with diverse scene structures and salient spatial cues. Therefore, even a small number of informative frames can sufficiently support spatial reasoning, while denser sampling introduces redundancy without significantly enhancing spatial understanding. This further validates the efficiency and robustness of \textsc{See\&Trek} in leveraging a compact, semantically diverse set of frames to enhance \textsc{MLLMs}.

\begin{table}[ht]
\centering
\caption{Ablation studies of different keyframe number $\mathcal{K}$ setting. Method ``-'' denotes the baseline \textsc{MLLM} \textsc{InternVL3-8B}. }
\resizebox{\linewidth}{!}{
\begin{tabular}{cc|c|cccccccc}
\toprule
\multirow{2}{*}{$\mathcal{K}$} & \multirow{2}{*}{Method} & \multirow{2}{*}{\textbf{Avg.}} & Obj. Count & Abs. Dist. & Obj. Size & Room Size & Rel. Dist. & Rel. Dir. & Route Plan & Appr. Order \\
& & & \multicolumn{4}{c}{\cellcolor{orange!10}\textbf{Numerical Answer}} & \multicolumn{4}{c}{\cellcolor{yellow!10}\textbf{Multiple-Choice Answer}} \\
 \midrule
  \multirow{2}{*}{8} & - & 40.2 & 67.8 & 32.2 & 44.5 & 41.8 & 42.8 & 37.7 & 26.3 & 28.3 \\
   & \textbf{\textsc{+See\&Trek}} & 43.2$_{\textcolor{red}{+3.0\%}}$ & 65.2$_{\textcolor{darkgreen}{-2.6\%}}$ & 32.9$_{\textcolor{red}{+0.7\%}}$ & 46.9$_{\textcolor{red}{+2.4\%}}$ & 46.7$_{\textcolor{red}{+4.9\%}}$ & 45.9$_{\textcolor{red}{+3.1\%}}$ & 40.2$_{\textcolor{red}{+2.5\%}}$ & 30.4$_{\textcolor{red}{+4.1\%}}$ & 37.4$_{\textcolor{red}{+9.1\%}}$ \\
    \midrule
    \multirow{2}{*}{12} & - & 41.8 & 67.8 & 32.2 & 44.5 & 41.8 & 42.8 & 37.7 & 26.3 & 28.3 \\
    & \textbf{\textsc{+See\&Trek}} & 43.6$_{\textcolor{red}{+1.8\%}}$ & 67.8$_{\textcolor{gray}{+0.0\%}}$ & 34.0$_{\textcolor{red}{+1.8\%}}$ & 43.4$_{\textcolor{darkgreen}{-1.1\%}}$ & 45.0$_{\textcolor{red}{+3.2\%}}$ & 43.9$_{\textcolor{red}{+1.1\%}}$ & 39.7$_{\textcolor{red}{+2.0\%}}$ & 26.3$_{\textcolor{gray}{+0.0\%}}$ & 34.3$_{\textcolor{red}{+6.0\%}}$ \\
    \midrule
    \multirow{2}{*}{16} & - & 42.4 & 69.3 & 33.9 & 43.1 & 47.3 & 44.8 & 38.7 & 26.8 & 35.3 \\
    & \textbf{\textsc{+See\&Trek}} & 44.0$_{\textcolor{red}{+1.6\%}}$ & 67.3$_{\textcolor{darkgreen}{-2.0\%}}$ & 34.6$_{\textcolor{red}{+0.7\%}}$ & 48.1$_{\textcolor{red}{+5.0\%}}$ & 47.9$_{\textcolor{red}{+0.6\%}}$ & 44.9$_{\textcolor{red}{+0.1\%}}$ & 39.7$_{\textcolor{red}{+1.0\%}}$ & 29.4$_{\textcolor{red}{+2.6\%}}$ & 39.6$_{\textcolor{red}{+4.3\%}}$ \\
    \midrule
    \multirow{2}{*}{24} & - & 43.2 & 69.6 & 35.8 & 44.3 & 45.9 & 49.2 & 39.1 & 27.8 & 33.7 \\
    & \textbf{\textsc{+See\&Trek}} & 44.1$_{\textcolor{red}{+0.9\%}}$ & 67.5$_{\textcolor{darkgreen}{-2.1\%}}$ & 34.5$_{\textcolor{darkgreen}{-1.3\%}}$ & 46.6$_{\textcolor{red}{+2.3\%}}$ & 49.0$_{\textcolor{red}{+3.1\%}}$ & 47.3$_{\textcolor{darkgreen}{-1.9\%}}$ & 39.0$_{\textcolor{darkgreen}{-0.1\%}}$ & 30.4$_{\textcolor{red}{+2.6\%}}$ & 38.7$_{\textcolor{red}{+5.0\%}}$ \\
    \midrule
    \multirow{2}{*}{32} & - & 43.1 & 69.3 & 34.9 & 41.6 & 49.4 & 46.9 & 38.9 & 29.9 & 33.9 \\
    & \textbf{\textsc{+See\&Trek}} & 43.7$_{\textcolor{red}{+0.6\%}}$ & 68.5$_{\textcolor{darkgreen}{-0.8\%}}$ & 34.3$_{\textcolor{darkgreen}{-0.6\%}}$ & 46.1$_{\textcolor{red}{+4.5\%}}$ & 48.1$_{\textcolor{darkgreen}{-1.3\%}}$ & 45.6$_{\textcolor{darkgreen}{-1.3\%}}$ & 41.5$_{\textcolor{red}{+2.6\%}}$ & 28.4$_{\textcolor{darkgreen}{-1.5\%}}$ & 36.7$_{\textcolor{red}{+2.8\%}}$ \\
    \bottomrule
\end{tabular}}
\label{tab:keyframe}
\vspace{-0.2cm}
\end{table}

\textbf{Motion Trajectory.} As mentioned in Section \ref{sec:Motion Reconstruction}, we also investigate performance impact of different part of input visual trajectories containing BEV $\mathbf{P}_\text{BEV}$ and 3D figures $\mathbf{P}_\text{3D}$. Table~\ref{tab:visualtraj} presents the ablation results evaluating the contribution of different visual trajectory components—BEV projections ($\mathbf{P}_\text{BEV}$) and 3D spatial figures ($\mathbf{P}_{\text{3D}}$) to the overall performance. Removing either component leads to a noticeable drop in average accuracy, confirming the complementary value of both trajectory views in enhancing spatial understanding. Specifically, excluding $\mathbf{P}_\text{BEV}$ results in slightly better performance than removing $\mathbf{P}_\text{3D}$, particularly in tasks requiring high-level route planning and temporal ordering, suggesting that BEV provides a more intuitive overview of the navigation path and spatial layout. Conversely, the 3D visualization offers critical depth and geometric cues, as evidenced by its impact on fine-grained spatial tasks such as "Obj. Size" and "Room Size." The results demonstrate that both $\mathbf{P}_\text{BEV}$ and $\mathbf{P}_\text{3D}$ are indispensable for supplying diverse and complementary motion context to the \textsc{MLLM}, enabling more robust reasoning across spatial tasks.

\begin{table}[ht]
\centering
\caption{Comparison results of different visual spatial methods containing \textsc{SoM}\cite{SoM}, \textsc{CC} \cite{CC} and our \textbf{\textsc{See\&Trek}}. It shows that \textbf{\textsc{See\&Trek}} obtains the SOTA performance in boosting the MLLMs spatial understanding.}
\resizebox{\linewidth}{!}{
\begin{tabular}{c|c|cccccccc}
    \toprule
\multirow{2}{*}{Method} & \multirow{2}{*}{\textbf{Avg.}} & Obj. Count & Abs. Dist. & Obj. Size & Room Size & Rel. Dist. & Rel. Dir. & Route Plan & Appr. Order \\
  &  & \multicolumn{4}{c}{\cellcolor{orange!10}\textbf{Numerical Answer}} & \multicolumn{4}{c}{\cellcolor{yellow!10}\textbf{Multiple-Choice Answer}} \\
 \midrule
\textsc{SoM} \cite{SoM} & 41.3 & 64.5 & 29.6 & 44.9 & 43.1 & 36.8 & 41.1 & 27.3 & 43.7 \\
\textsc{CC} \cite{CC} & 40.9 & 64.4 & 30.0 & 43.8 & 38.3 & 39.1 & 41.6 & 27.8 & 42.2 \\
\rowcolor{blue!10} \textbf{\textsc{See\&Trek}} & 43.2 & 65.2 & 32.9 & 46.9 & 46.7 & 45.9 & 40.2 & 30.4 & 37.4 \\
\bottomrule
\end{tabular}}
\label{tab:moreprompt}
\vspace{-0.2cm}
\end{table}

\subsection{Qualitative results}

\begin{figure}[t]
    \centering
    \includegraphics[width=\linewidth]{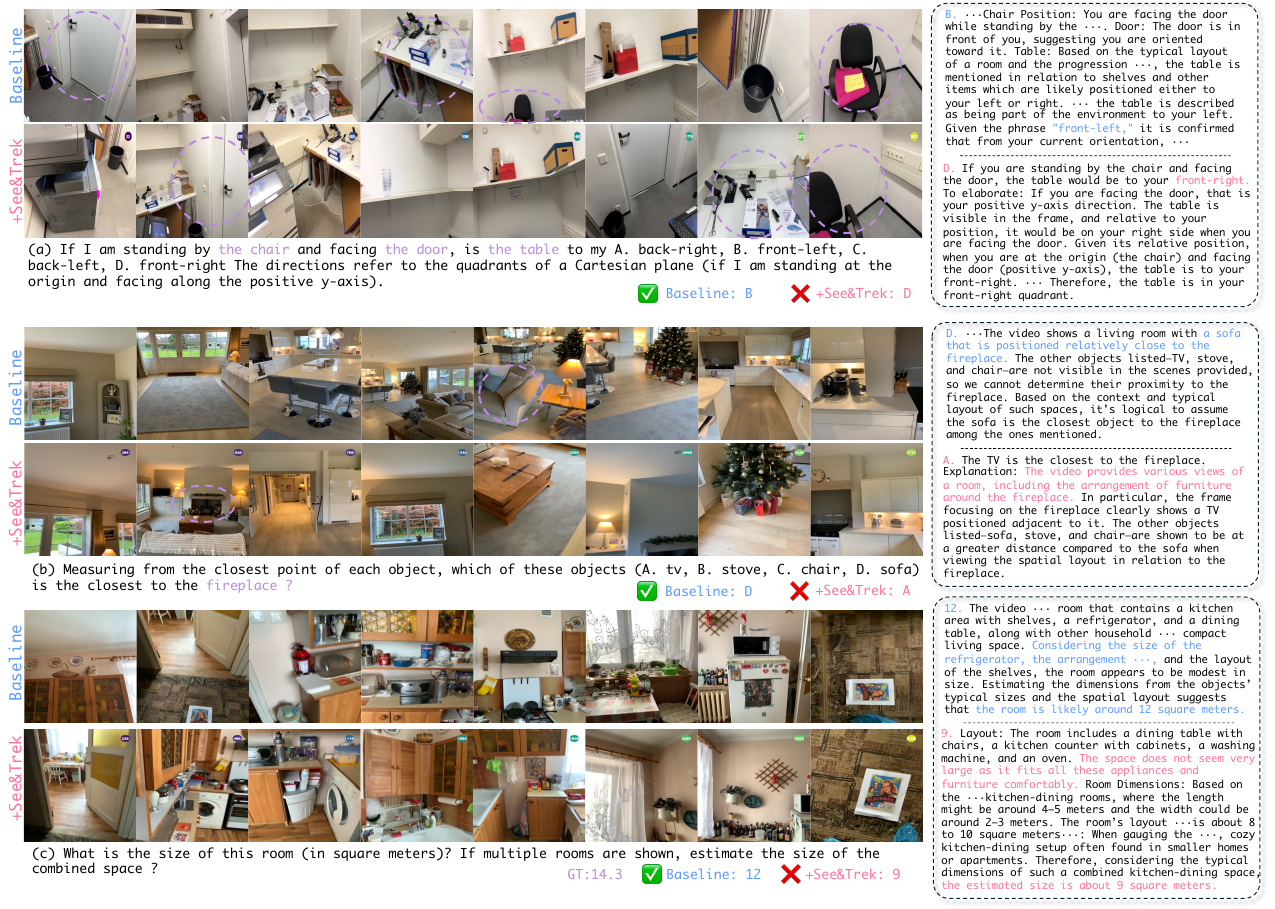}
    \vspace{-0.2cm}
    \caption{Illustration of several failure cases of \textbf{\textsc{See\&Trek}} compared to the baseline. It contains three spatial tasks: (a) route plan (b) object relative distance (c) room size. }
    \label{fig:failures}
\end{figure}

We give more illustration about how \textbf{\textsc{See\&Trek}} impacts the baseline inference as shown in Figure \ref{fig:results_1}/\ref{fig:results_2}/\ref{fig:results_3}. We also utilize the \textcolor{paperpurple}{purple circle} to highlight the objects mentioned in the question. For instance, in Figure \ref{fig:results_2}(e), we observe that \textbf{\textsc{See\&Trek}} can inspire more spatial reasoning in the baseline, such as more rational Chain of Thoughts (CoT) by incorporating more meaningful spatial information, leading to more accurate relative distance estimation. Besides, we also can conclude that the motion visual trajectories can improve the accuracy about the size estimation like Figure \ref{fig:results_1}(b) like \textcolor{paperred}{``given the visual cues.''} Furthermore, we can get a conclusion that the Maximum Semantic Richness Sampling (MSRS) can boost the richness of extracted frames from the given video as shown in Figure \ref{fig:results_1}(c) and \ref{fig:results_2}(d,e,f). Overall, the proposed \textbf{\textsc{See\&Trek}} nightlight its advantages in modeling egocentric movement and temporal coherence.

\subsection{Failure Cases} 
As shown in Figure \ref{fig:failures}, we also investigate the failure cases generate from \textbf{\textsc{See\&Trek}} with empirical study. We first conclude that our baseline model maybe confused by the provided spatial information in some domains. For instance, in Figure \ref{fig:failures}(a) or \ref{fig:failures}(b), we observe that even though existing all objects like \textcolor{paperpurple}{door} or \textcolor{paperpurple}{fireplace} from the query in the selected frames, the answer generated from \textbf{\textsc{See\&Trek}} is wrong. Besides, it also get the conclusion that choosing GPU-free YOLO as the perception model still does not have strong generalization ability and performs poorly in multi object and multi class scenes like Figure \ref{fig:failures}(c), which lacks more goals compared to the baseline. It motivates us to utilize more powerful dense perception model to extract higher semantic richness frames from the given spatial videos. 

\subsection{More Prompting Comparison} 

\textbf{\textsc{See\&Trek}} is the first training\&gpu-free method to improve the spatial understanding ability of \textsc{MLLMs}. Served as the prompting methods \cite{AttentionPrompting,mllmsknows,See,Mllmcansee}, we also explore other current visual prompting method \cite{3DAxisPrompt,SoM,CC} in other domains for comparison. We notice the \textsc{3DAxisPrompt} \cite{3DAxisPrompt} also can boost MLLMs spatial understanding which leverages the 3D coordinate axis and masks generated from the Segment Anything Model (SAM) \cite{SAM} to provide explicit geometric priors. Due to its lack of open-source code which makes it hard to review, we only discuss it here.
Then, we leverage Set of Mask (\textsc{SoM}) \cite{SoM} and Coarse Correspondence (\textsc{CC}) \cite{CC}, which enable the identification of objects—either with masks or numeric labels, by utilizing the capabilities of current vision-language models. Similar from the setting \cite{SoM,CC}, we also sample uniformly 8 frames from given videos and conduct these method to process. We still follow the above-mentioned setting to evaluate, \textit{e.g.} \textsc{InternVL3-8B} as baseline. As shown in Table \ref{tab:moreprompt}, our proposed \textbf{\textsc{See\&Trek}} consistently outperforms existing visual prompting methods such as \textsc{SoM} and \textsc{CC} across all spatial reasoning tasks. While \textsc{SoM} and \textsc{CC} primarily rely on static masks or coarse positional labels to identify object regions, they offer limited support for modeling inter-object relationships or capturing scene-level spatial structures. In contrast, \textbf{\textsc{See\&Trek}} leverages two key principles—visual diversity and motion reconstruction—to provide richer and more structured spatial cues. Specifically, the Maximum Semantic Richness Sampling strategy ensures that the input frames encapsulate diverse spatial layouts and semantic contexts, while the simulated visual trajectories explicitly encode relative positions and temporal continuity, which are essential for complex spatial reasoning tasks. Unlike prior approaches that depend on task-specific annotations or domain priors, \textbf{\textsc{See\&Trek}} operates in a fully training- and GPU-free manner, making it both lightweight and generalizable. This design enables \textsc{MLLMs} to develop a deeper and more coherent spatial understanding, as reflected in the consistent improvements across all evaluated tasks.

\section{Discussion: Trajectories Help MLLMs Construct Implicit Spatial Map}

\begin{figure}[t]
    \centering
    \includegraphics[width=\linewidth]{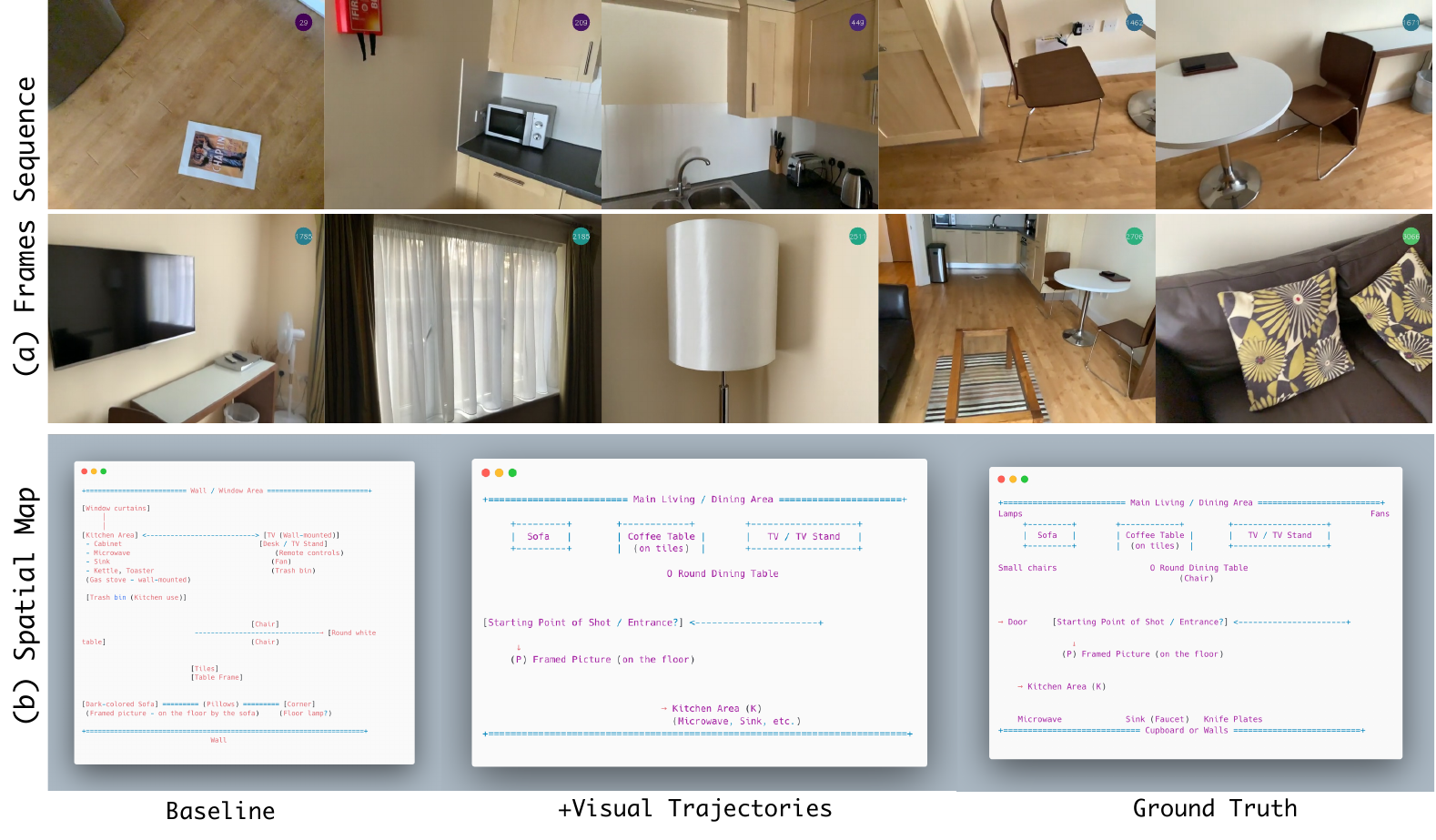}
    \vspace{-0.4cm}
    \caption{Illustration of the implicit impact of visual trajectories from {\textsc{See\&Trek}} for MLLMs internal inference. Here, we utilize \textsc{Gemini 2.5 Pro} as our baseline. It demonstrates that visual trajectories can help MLLMs construct better implicit spatial map, which improve different spatial task performance comprehensively.}
    \label{fig:implicit}
\end{figure}

As discussed in earlier sections, \textbf{\textsc{See\&Trek}} is designed to enhance the spatial understanding capabilities of \textsc{MLLMs} under limited visual input by preserving both spatial relationships and temporal coherence. While Maximum Semantic Richness Sampling effectively improves visual diversity, the influence of motion reconstruction on the model’s internal inference processes remains less understood. To investigate this, we move beyond explicit reasoning analyses (\textit{e.g.}, Figure~\ref{fig:results}) and propose a more challenging task: generating structured spatial maps in text form. This setting offers a more direct and interpretable view into the model's implicit spatial reasoning and internal representational mechanisms. As shown in Figure~\ref{fig:implicit}, the baseline model exhibits limited spatial coherence—although relevant entities are identified, their placements are often disorganized and misaligned with the actual room layout. In contrast, the model equipped with visual trajectories produces spatial maps that more closely resemble the ground truth. These outputs demonstrate more accurate object positioning (\textit{e.g.}, correctly situating the kitchen area and entrance) and better structural grouping (\textit{e.g.}, aligning the coffee table with the sofa and TV stand). The observed improvement underscores the critical role of motion-guided trajectories in reinforcing spatial continuity and layout reasoning. By temporally linking semantically rich keyframes, visual trajectories provide the model with stronger contextual cues, helping it infer object relationships and transitions across frames. In essence, these trajectories act as a soft inductive prior, enabling \textsc{MLLMs} to construct spatially consistent internal representations from fragmented visual observations.

\section{Limitations}

The idea core of \textbf{\textsc{See\&Trek}} is to compress various spatial information within the video and express it explicitly like vision. Although we have established a good prompt framework, its performance is still limited by the choice of external perception model. Our future work will further relax the condition assumptions, such as GPU-free, and explore the use of visual perception models with richer semantic knowledge to enhance MLLMs, \textit{e.g.} \textsc{Depth Anything}~\cite{Depthanything} or \textsc{GroundingDINO}~\cite{Groundingdino}.


\clearpage

\bibliographystyle{unsrt}
\small{\bibliography{reference}}

\begin{thebibliography}{10}

\bibitem{qwen25}
Shuai Bai, Keqin Chen, Xuejing Liu, Jialin Wang, Wenbin Ge, Sibo Song, Kai Dang, Peng Wang, Shijie Wang, Jun Tang, et~al.
\newblock Qwen2. 5-vl technical report.
\newblock {\em arXiv preprint arXiv:2502.13923}, 2025.

\bibitem{DeepseekR1}
Daya Guo, Dejian Yang, Haowei Zhang, Junxiao Song, Ruoyu Zhang, Runxin Xu, Qihao Zhu, Shirong Ma, Peiyi Wang, Xiao Bi, et~al.
\newblock Deepseek-r1: Incentivizing reasoning capability in llms via reinforcement learning.
\newblock {\em arXiv preprint arXiv:2501.12948}, 2025.

\bibitem{gpt}
Josh Achiam, Steven Adler, Sandhini Agarwal, Lama Ahmad, Ilge Akkaya, Florencia~Leoni Aleman, Diogo Almeida, Janko Altenschmidt, Sam Altman, Shyamal Anadkat, et~al.
\newblock Gpt-4 technical report.
\newblock {\em arXiv preprint arXiv:2303.08774}, 2023.

\bibitem{team2024gemini}
Gemini Team, Petko Georgiev, Ving~Ian Lei, Ryan Burnell, Libin Bai, Anmol Gulati, Garrett Tanzer, Damien Vincent, Zhufeng Pan, Shibo Wang, et~al.
\newblock Gemini 1.5: Unlocking multimodal understanding across millions of tokens of context.
\newblock {\em arXiv preprint arXiv:2403.05530}, 2024.

\bibitem{VideoMMMU}
Kairui Hu, Penghao Wu, Fanyi Pu, Wang Xiao, Yuanhan Zhang, Xiang Yue, Bo~Li, and Ziwei Liu.
\newblock Video-mmmu: Evaluating knowledge acquisition from multi-discipline professional videos.
\newblock {\em arXiv preprint arXiv:2501.13826}, 2025.

\bibitem{LLaVA-NeXT}
Haotian Liu, Chunyuan Li, Yuheng Li, Bo~Li, Yuanhan Zhang, Sheng Shen, and Yong~Jae Lee.
\newblock Llava-next: Improved reasoning, ocr, and world knowledge, January 2024.

\bibitem{liu2024nvila}
Zhijian Liu, Ligeng Zhu, Baifeng Shi, Zhuoyang Zhang, Yuming Lou, Shang Yang, Haocheng Xi, Shiyi Cao, Yuxian Gu, Dacheng Li, Xiuyu Li, Yunhao Fang, Yukang Chen, Cheng-Yu Hsieh, De-An Huang, An-Chieh Cheng, Vishwesh Nath, Jinyi Hu, Sifei Liu, Ranjay Krishna, Daguang Xu, Xiaolong Wang, Pavlo Molchanov, Jan Kautz, Hongxu Yin, Song Han, and Yao Lu.
\newblock Nvila: Efficient frontier visual language models, 2024.

\bibitem{Openvla}
Moo~Jin Kim, Karl Pertsch, Siddharth Karamcheti, Ted Xiao, Ashwin Balakrishna, Suraj Nair, Rafael Rafailov, Ethan Foster, Grace Lam, Pannag Sanketi, et~al.
\newblock Openvla: An open-source vision-language-action model.
\newblock {\em arXiv preprint arXiv:2406.09246}, 2024.

\bibitem{LLaKey}
Zheyi Zhao, Ying He, Fei Yu, Pengteng Li, Fan Zhuo, and Xilong Sun.
\newblock Llakey: Follow my basic action instructions to your next key state.
\newblock In {\em 2024 IEEE/RSJ International Conference on Intelligent Robots and Systems (IROS)}, pages 9604--9611. IEEE, 2024.

\bibitem{Drivegpt4}
Zhenhua Xu, Yujia Zhang, Enze Xie, Zhen Zhao, Yong Guo, Kwan-Yee~K Wong, Zhenguo Li, and Hengshuang Zhao.
\newblock Drivegpt4: Interpretable end-to-end autonomous driving via large language model.
\newblock {\em IEEE Robotics and Automation Letters}, 2024.

\bibitem{CotVla}
Qingqing Zhao, Yao Lu, Moo~Jin Kim, Zipeng Fu, Zhuoyang Zhang, Yecheng Wu, Zhaoshuo Li, Qianli Ma, Song Han, Chelsea Finn, et~al.
\newblock Cot-vla: Visual chain-of-thought reasoning for vision-language-action models.
\newblock {\em arXiv preprint arXiv:2503.22020}, 2025.

\bibitem{llava3d}
Chenming Zhu, Tai Wang, Wenwei Zhang, Jiangmiao Pang, and Xihui Liu.
\newblock Llava-3d: A simple yet effective pathway to empowering lmms with 3d-awareness.
\newblock {\em arXiv preprint arXiv:2409.18125}, 2024.

\bibitem{SpatialRGPT}
An-Chieh Cheng, Hongxu Yin, Yang Fu, Qiushan Guo, Ruihan Yang, Jan Kautz, Xiaolong Wang, and Sifei Liu.
\newblock Spatialrgpt: Grounded spatial reasoning in vision language models.
\newblock {\em arXiv preprint arXiv:2406.01584}, 2024.

\bibitem{Scanqa}
Daichi Azuma, Taiki Miyanishi, Shuhei Kurita, and Motoaki Kawanabe.
\newblock Scanqa: 3d question answering for spatial scene understanding.
\newblock In {\em proceedings of the IEEE/CVF conference on computer vision and pattern recognition}, pages 19129--19139, 2022.

\bibitem{Sqa3d}
Xiaojian Ma, Silong Yong, Zilong Zheng, Qing Li, Yitao Liang, Song-Chun Zhu, and Siyuan Huang.
\newblock Sqa3d: Situated question answering in 3d scenes.
\newblock {\em arXiv preprint arXiv:2210.07474}, 2022.

\bibitem{Mlvu}
Junjie Zhou, Yan Shu, Bo~Zhao, Boya Wu, Shitao Xiao, Xi~Yang, Yongping Xiong, Bo~Zhang, Tiejun Huang, and Zheng Liu.
\newblock Mlvu: A comprehensive benchmark for multi-task long video understanding.
\newblock {\em arXiv preprint arXiv:2406.04264}, 2024.

\bibitem{HierarQ}
Shehreen Azad, Vibhav Vineet, and Yogesh~Singh Rawat.
\newblock Hierarq: Task-aware hierarchical q-former for enhanced video understanding.
\newblock {\em arXiv preprint arXiv:2503.08585}, 2025.

\bibitem{breakingencoder}
Handong Li, Yiyuan Zhang, Longteng Guo, Xiangyu Yue, and Jing Liu.
\newblock Breaking the encoder barrier for seamless video-language understanding.
\newblock {\em arXiv preprint arXiv:2503.18422}, 2025.

\bibitem{QuoTA}
Yongdong Luo, Wang Chen, Xiawu Zheng, Weizhong Huang, Shukang Yin, Haojia Lin, Chaoyou Fu, Jinfa Huang, Jiayi Ji, Jiebo Luo, et~al.
\newblock Quota: Query-oriented token assignment via cot query decouple for long video comprehension.
\newblock {\em arXiv preprint arXiv:2503.08689}, 2025.

\bibitem{hybridtoken}
Zhihang Liu, Chen-Wei Xie, Pandeng Li, Liming Zhao, Longxiang Tang, Yun Zheng, Chuanbin Liu, and Hongtao Xie.
\newblock Hybrid-level instruction injection for video token compression in multi-modal large language models.
\newblock {\em arXiv preprint arXiv:2503.16036}, 2025.

\bibitem{yuan2025memory}
Huaying Yuan, Zheng Liu, Minhao Qin, Hongjin Qian, Y~Shu, Zhicheng Dou, and Ji-Rong Wen.
\newblock Memory-enhanced retrieval augmentation for long video understanding.
\newblock {\em arXiv preprint arXiv:2503.09149}, 2025.

\bibitem{infty-Video}
Saul Santos, Ant{\'o}nio Farinhas, Daniel~C McNamee, and Andr{\'e}~FT Martins.
\newblock infty-video: A training-free approach to long video understanding via continuous-time memory consolidation.
\newblock {\em arXiv preprint arXiv:2501.19098}, 2025.

\bibitem{ReWind}
Anxhelo Diko, Tinghuai Wang, Wassim Swaileh, Shiyan Sun, and Ioannis Patras.
\newblock Rewind: Understanding long videos with instructed learnable memory.
\newblock {\em arXiv preprint arXiv:2411.15556}, 2024.

\bibitem{ma2024drvideo}
Ziyu Ma, Chenhui Gou, Hengcan Shi, Bin Sun, Shutao Li, Hamid Rezatofighi, and Jianfei Cai.
\newblock Drvideo: Document retrieval based long video understanding.
\newblock {\em arXiv preprint arXiv:2406.12846}, 2024.

\bibitem{VideoRAG-l}
Xubin Ren, Lingrui Xu, Long Xia, Shuaiqiang Wang, Dawei Yin, and Chao Huang.
\newblock Videorag: Retrieval-augmented generation with extreme long-context videos.
\newblock {\em arXiv preprint arXiv:2502.01549}, 2025.

\bibitem{VideoRAG-s}
Yongdong Luo, Xiawu Zheng, Xiao Yang, Guilin Li, Haojia Lin, Jinfa Huang, Jiayi Ji, Fei Chao, Jiebo Luo, and Rongrong Ji.
\newblock Video-rag: Visually-aligned retrieval-augmented long video comprehension.
\newblock {\em arXiv preprint arXiv:2411.13093}, 2024.

\bibitem{Graphsvideo}
Zeyi Huang, Yuyang Ji, Xiaofang Wang, Nikhil Mehta, Tong Xiao, Donghyun Lee, Sigmund Vanvalkenburgh, Shengxin Zha, Bolin Lai, Licheng Yu, et~al.
\newblock Building a mind palace: Structuring environment-grounded semantic graphs for effective long video analysis with llms.
\newblock {\em arXiv preprint arXiv:2501.04336}, 2025.

\bibitem{Videotree}
Ziyang Wang, Shoubin Yu, Elias Stengel-Eskin, Jaehong Yoon, Feng Cheng, Gedas Bertasius, and Mohit Bansal.
\newblock Videotree: Adaptive tree-based video representation for llm reasoning on long videos.
\newblock {\em arXiv preprint arXiv:2405.19209}, 2024.

\bibitem{adaptiveframes}
Xi~Tang, Jihao Qiu, Lingxi Xie, Yunjie Tian, Jianbin Jiao, and Qixiang Ye.
\newblock Adaptive keyframe sampling for long video understanding.
\newblock {\em arXiv preprint arXiv:2502.21271}, 2025.

\bibitem{agentic-keyframes}
Sunqi Fan, Meng-Hao Guo, and Shuojin Yang.
\newblock Agentic keyframe search for video question answering.
\newblock {\em arXiv preprint arXiv:2503.16032}, 2025.

\bibitem{GenerativeFrameSampler}
Linli Yao, Haoning Wu, Kun Ouyang, Yuanxing Zhang, Caiming Xiong, Bei Chen, Xu~Sun, and Junnan Li.
\newblock Generative frame sampler for long video understanding.
\newblock {\em arXiv preprint arXiv:2503.09146}, 2025.

\bibitem{LogicinFrames}
Weiyu Guo, Ziyang Chen, Shaoguang Wang, Jianxiang He, Yijie Xu, Jinhui Ye, Ying Sun, and Hui Xiong.
\newblock Logic-in-frames: Dynamic keyframe search via visual semantic-logical verification for long video understanding.
\newblock {\em arXiv preprint arXiv:2503.13139}, 2025.

\bibitem{Selection}
Kai Hu, Feng Gao, Xiaohan Nie, Peng Zhou, Son Tran, Tal Neiman, Lingyun Wang, Mubarak Shah, Raffay Hamid, Bing Yin, et~al.
\newblock M-llm based video frame selection for efficient video understanding.
\newblock {\em arXiv preprint arXiv:2502.19680}, 2025.

\bibitem{drivesurvey}
Can Cui, Yunsheng Ma, Xu~Cao, Wenqian Ye, Yang Zhou, Kaizhao Liang, Jintai Chen, Juanwu Lu, Zichong Yang, Kuei-Da Liao, et~al.
\newblock A survey on multimodal large language models for autonomous driving.
\newblock In {\em Proceedings of the IEEE/CVF Winter Conference on Applications of Computer Vision}, pages 958--979, 2024.

\bibitem{GPT4Scene}
Zhangyang Qi, Zhixiong Zhang, Ye~Fang, Jiaqi Wang, and Hengshuang Zhao.
\newblock Gpt4scene: Understand 3d scenes from videos with vision-language models.
\newblock {\em arXiv preprint arXiv:2501.01428}, 2025.

\bibitem{STVLM}
Dohwan Ko, Sihyeon Kim, Yumin Suh, Minseo Yoon, Manmohan Chandraker, Hyunwoo~J Kim, et~al.
\newblock St-vlm: Kinematic instruction tuning for spatio-temporal reasoning in vision-language models.
\newblock {\em arXiv preprint arXiv:2503.19355}, 2025.

\bibitem{kimi}
Kimi Team, Angang Du, Bohong Yin, Bowei Xing, Bowen Qu, Bowen Wang, Cheng Chen, Chenlin Zhang, Chenzhuang Du, Chu Wei, et~al.
\newblock Kimi-vl technical report.
\newblock {\em arXiv preprint arXiv:2504.07491}, 2025.

\bibitem{vsibench}
Jihan Yang, Shusheng Yang, Anjali~W Gupta, Rilyn Han, Li~Fei-Fei, and Saining Xie.
\newblock Thinking in space: How multimodal large language models see, remember, and recall spaces.
\newblock {\em arXiv preprint arXiv:2412.14171}, 2024.

\bibitem{STIBench}
Yun Li, Yiming Zhang, Tao Lin, XiangRui Liu, Wenxiao Cai, Zheng Liu, and Bo~Zhao.
\newblock Sti-bench: Are mllms ready for precise spatial-temporal world understanding?
\newblock {\em arXiv preprint arXiv:2503.23765}, 2025.

\bibitem{ultralytics2023yolov8}
Glenn Jocher, Ayush Chaurasia, and Jing Qiu.
\newblock Ultralytics yolov8.
\newblock https://github.com/ultralytics/ultralytics, 2023.

\bibitem{RANSAC}
Ond{\v{r}}ej Chum, Ji{\v{r}}{\'\i} Matas, and Josef Kittler.
\newblock Locally optimized ransac.
\newblock In {\em Joint pattern recognition symposium}, pages 236--243. Springer, 2003.

\bibitem{OrbSlam3}
Carlos Campos, Richard Elvira, Juan J~G{\'o}mez Rodr{\'\i}guez, Jos{\'e}~MM Montiel, and Juan~D Tard{\'o}s.
\newblock Orb-slam3: An accurate open-source library for visual, visual--inertial, and multimap slam.
\newblock {\em IEEE transactions on robotics}, 37(6):1874--1890, 2021.

\bibitem{OrbSlam}
Raul Mur-Artal, Jose Maria~Martinez Montiel, and Juan~D Tardos.
\newblock Orb-slam: A versatile and accurate monocular slam system.
\newblock {\em IEEE transactions on robotics}, 31(5):1147--1163, 2015.

\bibitem{huber2011robust}
Peter~J Huber and Elvezio~M Ronchetti.
\newblock {\em Robust statistics}.
\newblock John Wiley \& Sons, 2011.

\bibitem{Scannet}
Angela Dai, Angel~X Chang, Manolis Savva, Maciej Halber, Thomas Funkhouser, and Matthias Nie{\ss}ner.
\newblock Scannet: Richly-annotated 3d reconstructions of indoor scenes.
\newblock In {\em Proceedings of the IEEE conference on computer vision and pattern recognition}, pages 5828--5839, 2017.

\bibitem{yeshwanth2023scannet++}
Chandan Yeshwanth, Yueh-Cheng Liu, Matthias Nie{\ss}ner, and Angela Dai.
\newblock Scannet++: A high-fidelity dataset of 3d indoor scenes.
\newblock In {\em Proceedings of the IEEE/CVF International Conference on Computer Vision}, pages 12--22, 2023.

\bibitem{Arkitscenes}
Gilad Baruch, Zhuoyuan Chen, Afshin Dehghan, Tal Dimry, Yuri Feigin, Peter Fu, Thomas Gebauer, Brandon Joffe, Daniel Kurz, Arik Schwartz, et~al.
\newblock Arkitscenes: A diverse real-world dataset for 3d indoor scene understanding using mobile rgb-d data.
\newblock {\em arXiv preprint arXiv:2111.08897}, 2021.

\bibitem{zhu2025internvl3}
Jinguo Zhu, Weiyun Wang, Zhe Chen, Zhaoyang Liu, Shenglong Ye, Lixin Gu, Yuchen Duan, Hao Tian, Weijie Su, Jie Shao, et~al.
\newblock Internvl3: Exploring advanced training and test-time recipes for open-source multimodal models.
\newblock {\em arXiv preprint arXiv:2504.10479}, 2025.

\bibitem{Llava-onevision}
Bo~Li, Yuanhan Zhang, Dong Guo, Renrui Zhang, Feng Li, Hao Zhang, Kaichen Zhang, Peiyuan Zhang, Yanwei Li, Ziwei Liu, et~al.
\newblock Llava-onevision: Easy visual task transfer.
\newblock {\em arXiv preprint arXiv:2408.03326}, 2024.

\bibitem{Omni6dpose}
Jiyao Zhang, Weiyao Huang, Bo~Peng, Mingdong Wu, Fei Hu, Zijian Chen, Bo~Zhao, and Hao Dong.
\newblock Omni6dpose: A benchmark and model for universal 6d object pose estimation and tracking.
\newblock In {\em European Conference on Computer Vision}, pages 199--216. Springer, 2024.

\bibitem{Waymo}
Pei Sun, Henrik Kretzschmar, Xerxes Dotiwalla, Aurelien Chouard, Vijaysai Patnaik, Paul Tsui, James Guo, Yin Zhou, Yuning Chai, Benjamin Caine, et~al.
\newblock Scalability in perception for autonomous driving: Waymo open dataset.
\newblock In {\em Proceedings of the IEEE/CVF conference on computer vision and pattern recognition}, pages 2446--2454, 2020.

\bibitem{SoM}
Jianwei Yang, Hao Zhang, Feng Li, Xueyan Zou, Chunyuan Li, and Jianfeng Gao.
\newblock Set-of-mark prompting unleashes extraordinary visual grounding in gpt-4v.
\newblock {\em arXiv preprint arXiv:2310.11441}, 2023.

\bibitem{CC}
Benlin Liu, Yuhao Dong, Yiqin Wang, Yongming Rao, Yansong Tang, Wei-Chiu Ma, and Ranjay Krishna.
\newblock Coarse correspondence elicit 3d spacetime understanding in multimodal language model.
\newblock {\em arXiv preprint arXiv:2408.00754}, 2024.

\bibitem{AttentionPrompting}
Runpeng Yu, Weihao Yu, and Xinchao Wang.
\newblock Attention prompting on image for large vision-language models.
\newblock In {\em European Conference on Computer Vision}, pages 251--268. Springer, 2024.

\bibitem{mllmsknows}
Jiarui Zhang, Mahyar Khayatkhoei, Prateek Chhikara, and Filip Ilievski.
\newblock Mllms know where to look: Training-free perception of small visual details with multimodal llms.
\newblock {\em arXiv preprint arXiv:2502.17422}, 2025.

\bibitem{See}
Seil Kang, Jinyeong Kim, Junhyeok Kim, and Seong~Jae Hwang.
\newblock See what you are told: Visual attention sink in large multimodal models.
\newblock {\em arXiv preprint arXiv:2503.03321}, 2025.

\bibitem{Mllmcansee}
Chenxi Wang, Xiang Chen, Ningyu Zhang, Bozhong Tian, Haoming Xu, Shumin Deng, and Huajun Chen.
\newblock Mllm can see? dynamic correction decoding for hallucination mitigation.
\newblock {\em arXiv preprint arXiv:2410.11779}, 2024.

\bibitem{3DAxisPrompt}
Dingning Liu, Cheng Wang, Peng Gao, Renrui Zhang, Xinzhu Ma, Yuan Meng, and Zhihui Wang.
\newblock 3daxisprompt: Promoting the 3d grounding and reasoning in gpt-4o.
\newblock {\em Neurocomputing}, 637:130072, 2025.

\bibitem{SAM}
Alexander Kirillov, Eric Mintun, Nikhila Ravi, Hanzi Mao, Chloe Rolland, Laura Gustafson, Tete Xiao, Spencer Whitehead, Alexander~C Berg, Wan-Yen Lo, et~al.
\newblock Segment anything.
\newblock In {\em Proceedings of the IEEE/CVF international conference on computer vision}, pages 4015--4026, 2023.

\bibitem{Depthanything}
Lihe Yang, Bingyi Kang, Zilong Huang, Zhen Zhao, Xiaogang Xu, Jiashi Feng, and Hengshuang Zhao.
\newblock Depthanything v2.
\newblock {\em Advances in Neural Information Processing Systems}, 37:21875--21911, 2024.

\bibitem{Groundingdino}
Tianhe Ren, Qing Jiang, Shilong Liu, Zhaoyang Zeng, Wenlong Liu, Han Gao, Hongjie Huang, Zhengyu Ma, Xiaoke Jiang, Yihao Chen, et~al.
\newblock Grounding dino 1.5: Advance the" edge" of open-set object detection.
\newblock {\em arXiv preprint arXiv:2405.10300}, 2024.

\end{thebibliography}

\end{document}